\documentclass[10pt,twocolumn]{article}
\pdfoutput=1
\usepackage{url}
\usepackage{cvpr}
\usepackage{times}
\usepackage[colorlinks=true, urlcolor=blue, pdfborder={0 0 0}]{hyperref}

\usepackage[english]{babel}
\usepackage{url}
\usepackage{array}
\usepackage{ textcomp }
\usepackage{amsmath}
\usepackage{amssymb}
\usepackage{wasysym}
\usepackage{graphicx}
\usepackage{subfigure}
\usepackage{booktabs}
\usepackage{threeparttable}
\usepackage{longtable}
\usepackage{colortbl}  
\usepackage{pbox}
\usepackage{textcomp} 
\usepackage{nicefrac} 
\usepackage{stfloats}

\usepackage{adjustbox}

\usepackage{xspace}
\renewcommand*{\eg}{e.g.\@\xspace}

\newcommand*{\ea}{et al.\@\xspace}

\usepackage{xcolor}

\newcommand\largeheight{0.09\columnwidth}
\newcommand\smallheight{0.07\columnwidth}
\newcommand\stretchy{0.2}

\usepackage{etoolbox}
\patchcmd{\thebibliography}{\section*{\refname}}{}{}{}

\makeatletter
\let\oldlt\longtable
\let\endoldlt\endlongtable
\def\longtable{\@ifnextchar[\longtable@i \longtable@ii}
\def\longtable@i[#1]{\begin{figure}[t]
\onecolumn
\begin{minipage}{0.5\textwidth}
\oldlt[#1]
}
\def\longtable@ii{\begin{figure}[t]
\onecolumn
\begin{minipage}{0.5\textwidth}
\oldlt
}
\def\endlongtable{\endoldlt
\end{minipage}
\twocolumn
\end{figure}}
\makeatother

\graphicspath{{figs/}}

\cvprfinalcopy
\def\cvprPaperID{5} 

\title{RGBD Datasets: Past, Present and Future}

\author{Michael Firman \\ University College London \\ \url{http://www.michaelfirman.co.uk/RGBDdatasets/} }

\date{\today}

\begin{document}
\maketitle



\maketitle

\begin{abstract}
  Since the launch of the Microsoft Kinect, scores of RGBD datasets have been released. These have propelled advances in areas from reconstruction to gesture recognition. In this paper we explore the field, reviewing datasets across eight categories: semantics, object pose estimation, camera tracking, scene reconstruction, object tracking, human actions, faces and identification. By extracting relevant information in each category we help researchers to find appropriate data for their needs, and we consider which datasets have succeeded in driving computer vision forward and why.

  Finally, we examine the future of RGBD datasets.
  We identify key areas which are currently underexplored, and suggest that future directions may include synthetic data and dense reconstructions of static and dynamic scenes.


\end{abstract}

\section{Introduction}





\begin{figure}[t]
  \footnotesize
  \begin{tabular}{m{3.5cm} b{4cm}}
  \includegraphics[width=0.43\columnwidth]{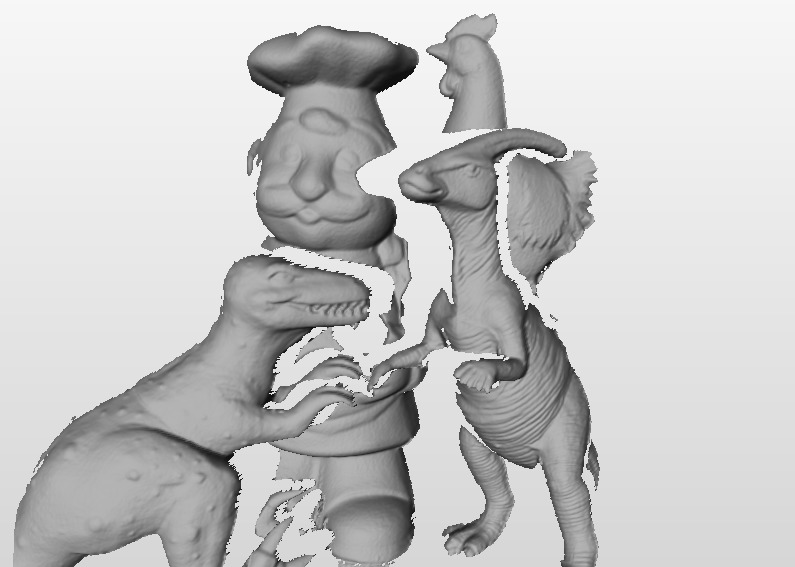}
  &
  \parbox[c]{4cm}{(a) \textbf{Past} \vspace{4pt} \\
  Before the Microsoft Kinect, most depth datasets were small and captured in the laboratory. \vspace{4pt} \\
  Image from \cite{mian-pami-2006}}
  \\
  \includegraphics[width=0.43\columnwidth]{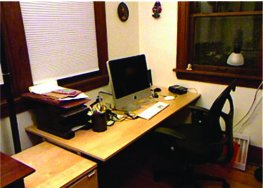}
  &
  \parbox[c]{4cm}{(b) \textbf{Present}   \vspace{4pt} \\
  We now enjoy RGBD data from dynamic and static scenes from the real world, with a range of labeling and capture conditions.  \vspace{4pt} \\
  Image from \cite{silberman-eccv-2012}
  }
  \\
  \includegraphics[width=0.43\columnwidth]{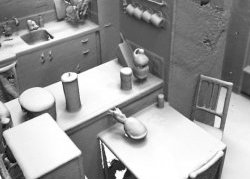}
  &
  \parbox[c]{4cm}{(c) \textbf{Future}  \vspace{4pt} \\
  We can anticipate scans of static and dynamic scenes as fused geometry, exploiting improvements in reconstruction algorithms. \vspace{4pt} \\
  Image from \cite{choi-cvpr-2015}}
  \\
  \end{tabular}
  \caption{The past, present and future of RGBD datasets.}
  \vspace{-4pt}
\label{fig:header2}
\end{figure}

Before the Microsoft Kinect was launched in November 2010, collecting images with a depth channel was a cumbersome and expensive task.
Researchers built custom active stereo setups \cite{breitenstein-cvpr-2008} and made use of 3D scanners costing tens of thousands of dollars \cite{mian-pami-2006, casia-minolta-2008}.
Many of these early datasets captured static images of objects in isolation, as the sensors used did not transport easily (Fig \ref{fig:header2}a).

\begin{figure}
    \includegraphics[width=1.0\columnwidth]{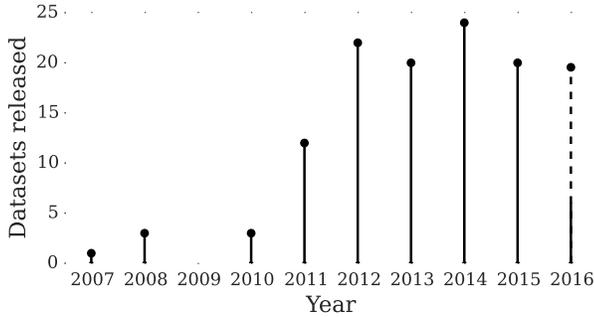}
    \caption{Our estimate of the number of depth datasets released each year, where projected releases in 2016 are shown as a dashed line. The Kinect was first released in November 2010. }
    \label{fig:datasets_per_year}
\end{figure}

Early Kinect datasets also focused on static images, often of single objects or small scenes.
As the field matures we see research being put to effect in creating larger and more ambitious RGBD datasets, and the quantity released each year shows no sign of decreasing (Figure \ref{fig:datasets_per_year}).
Semantic labels have been propagated through videos \cite{xiao-iccv-2013}, dense reconstruction has been exploited to capture the surfaces of whole objects \cite{choi-arxiv-2016} and generative scene algorithms have been used to create plausible synthetic data \cite{handa-arxiv-2015}.
We also see new labels applied to existing data \cite{guo-iccv-2013} and previous releases being recompiled into new offerings \cite{song-cvpr-2015}.




In spite of the current availability of sensors, though, collecting RGBD data is still not trivial.
Researchers using the Kinect have built battery devices \cite{silberman-eccv-2012, song-cvpr-2015}, written drivers \cite{song-cvpr-2015} and developed custom data formats \cite{freeman-arxiv-2015}.
Publicly available RGBD datasets can, at the most basic level, remove the need to repeat data capture.
More importantly, they provide transparency in the presentation of results and allow for scores to be compared on the same data by different researchers.
This in turn can drive competition for better-performing algorithms.
Finally, a dataset can help draw research towards previously under-explored directions.



\begin{table*}[bp]
\centering
\footnotesize
\renewcommand\arraystretch{\stretchy}
\caption{Datasets capturing single objects in isolation}
\label{tab:turntable}
\begin{threeparttable}
\begin{tabular}{m{1cm} m{5cm} m{4cm} c c c}
\toprule
&
& \textbf{Device\tnote{a}}
& \textbf{\# objects}
& \textbf{Camera pose?\tnote{b}}
& \textbf{Year} \\
\midrule
\includegraphics[height=\largeheight]{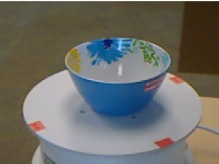} &
RGBD Object Dataset \cite{lai-icra-2011} &
Kinect v1 &
300 &
- &
'11 \\
\includegraphics[height=\largeheight]{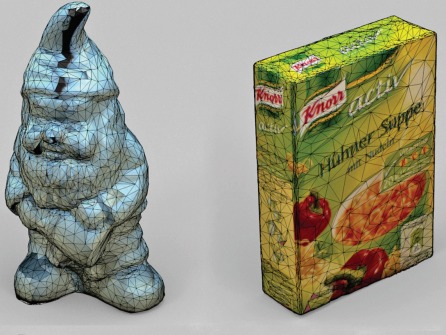} &
KIT object database \cite{kasper-ijrr-2012} &
Minolta Vi-900 and stereo pair &
$>$100 &
\checkmark \checkmark &
'12 \\
\includegraphics[height=\largeheight]{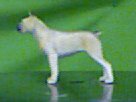} &
A dataset of Kinect-based 3D scans \cite{doumanoglou-ivmsp-2013} &
Kinect and Minolta Vi-900 &
59 &
\checkmark \checkmark &
'13 \\
\includegraphics[height=\largeheight]{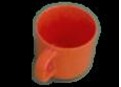} &
MV-RED \cite{liu-is-2014} &
Kinect v1 &
505 &
- &
'14 \\
\includegraphics[height=\largeheight]{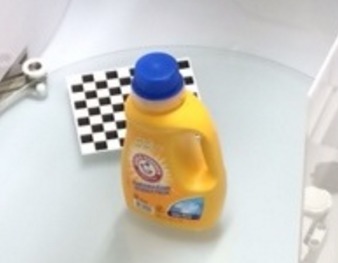} &
BigBIRD dataset \cite{singh-icra-2014} &
Asus Xtion Pro, DSLR &
125 &
\checkmark \checkmark &
'14 \\
\includegraphics[height=\largeheight]{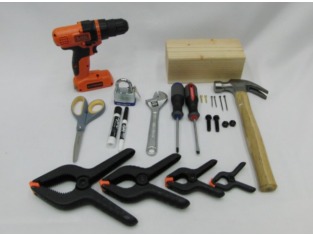} &
YCB Object and Model Set\tnote{c} ~\cite{calli-icar-2015} &
Asus Xtion Pro, DSLR &
88 &
\checkmark \checkmark &
'15 \\
\includegraphics[height=\largeheight]{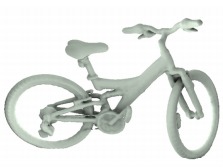} &
A large dataset of object scans \cite{choi-arxiv-2016} &
PrimeSense Carmine &
$>$10,000 &
\checkmark &
'16 \\
\bottomrule
\end{tabular}
\begin{tablenotes}
  \item[a] The Kinect v1, Asus Xtion Pro and PrimeSense Carmine have almost identical internals and can be considered to give equivalent data.
  \item[b] \checkmark = camera pose computed from RGBD data; \checkmark \checkmark = camera pose from calibration system.
  \item[c] Captured using the same turntable setup as the BigBIRD dataset.
\end{tablenotes}
\end{threeparttable}
\end{table*}

Our primary contribution is to give a snapshot of public RGBD datasets, allowing  researchers to easily select data appropriate for their needs (Section \ref{sec:present}).
We are more comprehensive than earlier efforts, describing 101 datasets compared with the 14 in \cite{berger-arxiv-2013}, 19 in \cite{guo-iciea-2014}\footnote{\cite{guo-iciea-2014} references more than 19 datasets, but most are not RGBD} and the 44 action datasets in \cite{zhang-arxiv-2016}. 
We secondly identify areas where there is opportunity for new data to facilitate novel areas of research (Section \ref{sec:future}).
We hypothesize that we can expect datasets to continue to move away from single images, to dense reconstructions of static and dynamic scenes (Figure \ref{fig:header2}c).

\section{State-of-the-art in RGBD datasets}
\label{sec:present}
Here we review state-of-the-art datasets across eight categories.
Some fall into more than one category, and the difference between categories depends as much on the labeling as it does the image content.

We include datasets which have been captured with an active capture devices such as time-of-flight or structured light, but exclude data from passive stereo.
We also exclude Lidar datasets, focusing instead on data from the separate world of commodity depth capture. 
Following the mantra that `data is cheap, information is expensive', we focus on data which has some form of human labeling applied. 
We exclude very small datasets, and those which have been produced mainly to demonstrate an acquisition method. 

With these exceptions, we aim to be comprehensive and correct.
Please flag omissions and errors to \url{m.firman@cs.ucl.ac.uk} so this document can be updated.
We also maintain a web-based version\footnote{\url{http://www.michaelfirman.co.uk/RGBDdatasets/}}.


We first look at datasets of objects in isolation, before moving on to datasets for camera tracking, scene reconstruction and then datasets where the pose of objects is to be inferred. Semantic, and then tracking datasets come next, before videos for action and gesture recognition. We finish with two more categories involving humans: faces and identity recognition.

\subsection{Objects in isolation}
\begin{table*}[bp]
\centering
\footnotesize
\renewcommand\arraystretch{\stretchy}
\caption{Datasets for camera pose and scene reconstruction}
\label{tab:camerapose}
\begin{threeparttable}
\newcolumntype{L}{>{\centering\arraybackslash}m{0.8cm}}
\newcolumntype{T}{>{\centering\arraybackslash}m{1.8cm}}
\begin{tabular}{m{0.8cm} >{\raggedright}m{4.1cm} m{1.5cm} c L T m{3.5cm} c}
\toprule
&
& \textbf{Device\tnote{a}}
& \textbf{\# videos}
& \textbf{Camera pose\tnote{b}}
& \textbf{Ground truth surface}
& \textbf{Notes}
& \textbf{Year} \\
\midrule
%
\includegraphics[height=\largeheight]{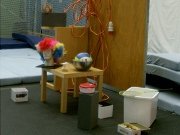} &
IROS 2011 Paper Kinect Dataset \cite{pomerleau-iros-2011} &
Kinect v1 &
27 &
\checkmark \checkmark &
&
- &
'11 \\
\includegraphics[height=\largeheight]{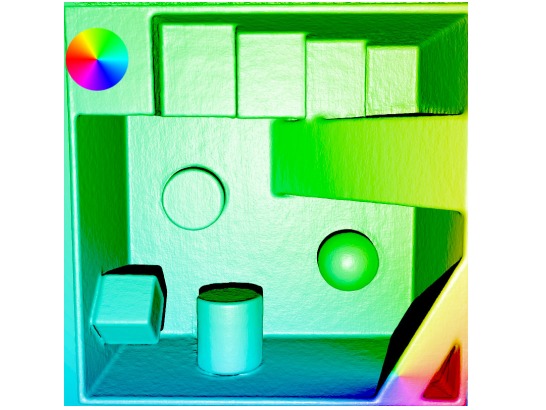} &
KinectFusion for Ground Truth \cite{meister-iros-2012} &
Kinect v1 &
 &
\checkmark  &
\checkmark &
Lidar surface ground truth for some scenes &
'12 \\
\includegraphics[height=\largeheight]{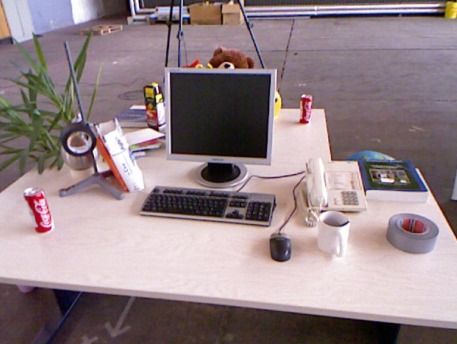} &
TUM benchmark \cite{sturm-iros-2012} &
Kinect v1 &
47 &
\checkmark \checkmark &
&
- &
'12 \\
\includegraphics[height=\largeheight]{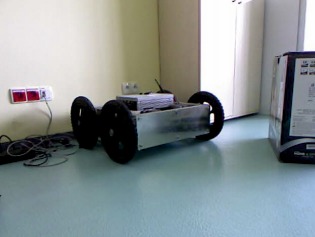} &
Indoor RGB-D Dataset \cite{schmidt-acivs-2013} &
Kinect v1 &
4 &
\checkmark \checkmark &
&
Collected from a robot &
'13 \\
\includegraphics[height=\largeheight]{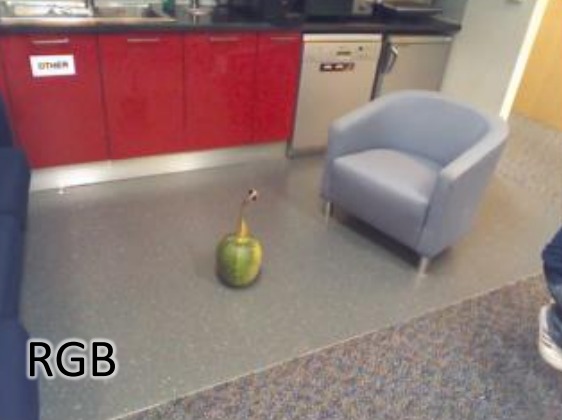} &
Microsoft 7-scenes \cite{shotton-cvpr-2013} &
Kinect v1 &
$>$14 &
\checkmark  &
 &
Designed for camera relocalization tasks &
'13 \\
\includegraphics[height=\largeheight]{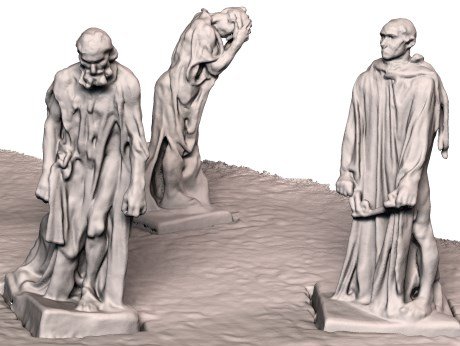} &
Robust Reconstruction Datasets \cite{zhou-siggraph-2013} &
Asus Xtion Pro &
8 &
\checkmark  &
 &
- &
'13 \\
\includegraphics[height=\largeheight]{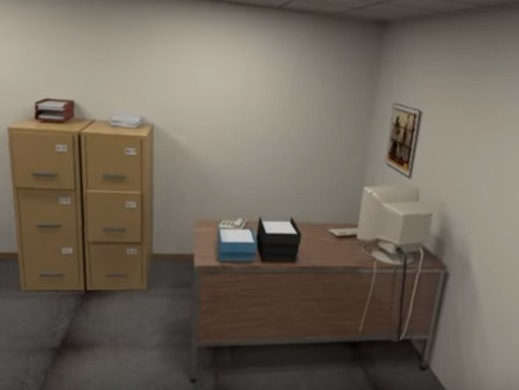} &
ICL-NUIM Dataset \cite{handa-icra-2014} &
Synthetic &
8 &
\checkmark \checkmark &
\checkmark &
Camera paths from \cite{choi-cvpr-2015} allow for reconstruction evaluation &
'14 \\
\includegraphics[height=\largeheight]{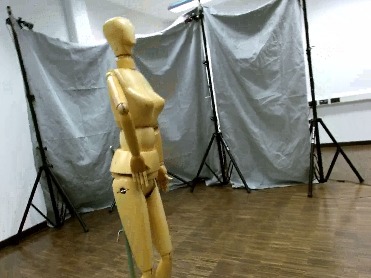} &
CoRBS Dataset \cite{wasenmueller-wacv-2016} &
Kinect v2 &
20 &
\checkmark \checkmark &
\checkmark &
Surface ground truth from fixed structured light scanner &
'16 \\
\includegraphics[height=\largeheight]{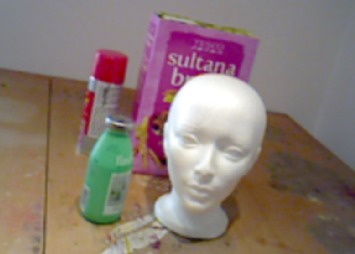} &
Voxel Occupancy Prediction \cite{firman-cvpr-2016} &
Asus Xtion Pro &
90 &
\checkmark  &
 &
Densely captures full visible surface &
'16 \\
\bottomrule
\end{tabular}
\begin{tablenotes}
  \item[a] The Kinect v1, Asus Xtion Pro and PrimeSense Carmine have almost identical internals and can be considered to give equivalent data.
  \item[b] \checkmark = approximated camera pose from Kinect tracking. \checkmark \checkmark = ground truth camera pose from external system.
\end{tablenotes}
\end{threeparttable}
\end{table*}

Following earlier stereo setups such as \cite{moreels-iccv-2005}, RGBD turntable datasets offer multiple unoccluded views of the same object from different angles (Table \ref{tab:turntable}).

The 2011 RGB-D Object Dataset \cite{lai-icra-2011} is a well-used dataset with 300 objects, but does not contain accurate camera poses. This was rectified by more recent datasets such as BigBIRD \cite{singh-icra-2014}. While a smaller dataset, BigBIRD is captured with calibrated Kinects and DSLRs.

Turntable datasets have been exploited in `natural' scenes for tasks such as object detection \cite{lai-icra-2012} and discovery \cite{firman-iros-2013}.
In many ways, though, they are limited by their deviation from real-world data.
Without occlusion, lighting changes or varying distances to objects these datasets sit in a different domain to the real-world scenes which we ultimately aim to understand.

Choi \ea \cite{choi-arxiv-2016} exploit improvements in camera tracking to form a dataset of individual objects scanned in the real world. With 10,000 items ranging in size from books to cars, this is the largest dataset of real-life objects by two orders of magnitude.

\subsection{Camera tracking and scene reconstruction}
Arguably some of the main advances brought by consumer depth cameras have been in camera tracking and dense reconstruction.
Ground truth camera poses are necessary to validate these algorithms, and these are difficult to acquire as they require external hardware.

For \textbf{camera tracking}, the TUM benchmark \cite{sturm-iros-2012} has become a de-facto standard for evaluation, with ground truth data from a motion tracking system and a range of scenes and camera motions.
We summarize this and similar datasets in Table~\ref{tab:camerapose}.

\begin{table*}[bp]
\centering
\footnotesize
\renewcommand\arraystretch{\stretchy}
\caption{Datasets for object pose estimation}
\label{tab:objectpose}
\begin{threeparttable}
\begin{tabular}{m{0.9cm} m{4cm} c c c m{4.5cm} c}
\toprule
&
& \textbf{Device}
& \textbf{\# objects}
& \textbf{\# frames}
& \textbf{Notes}
& \textbf{Year} \\
\midrule
\includegraphics[height=\largeheight]{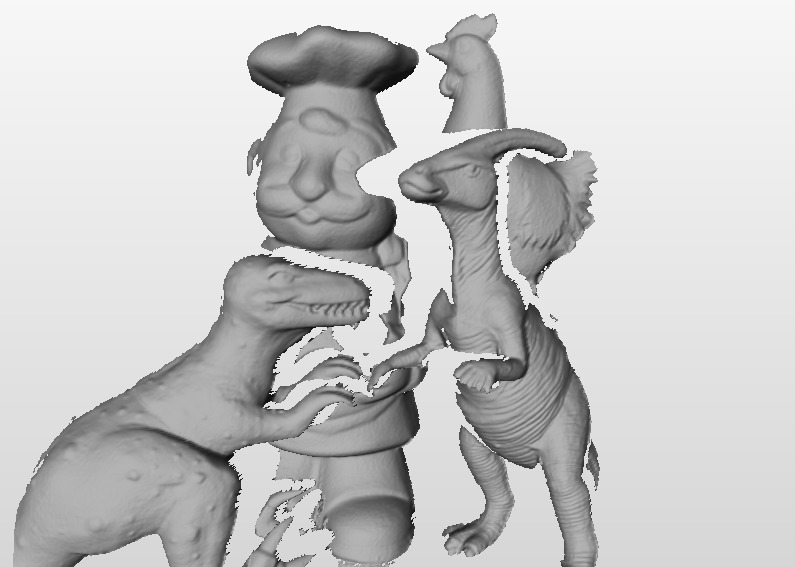} &
Cluttered scenes dataset \cite{mian-pami-2006} &
Minolta Vivid 910 &
5 &
48 &
Manual ground truth alignment &
'06
\\
\includegraphics[height=\largeheight]{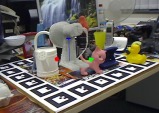} &
LINEMOD RGBD dataset \cite{hinterstoisser-accv-2012} &
Kinect v1 &
15 &
$>$18,000 &
Ground truth from calibration board  &
'12 \\
\includegraphics[height=\largeheight]{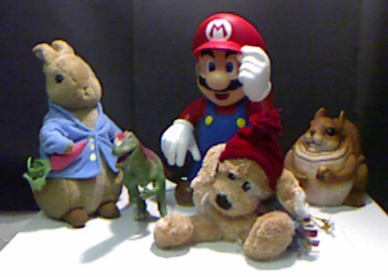} &
SHOT dataset \cite{salti-cviu-2014} &
Kinect v1 &
6 &
16 &
- &
'14 \\
\includegraphics[height=\largeheight]{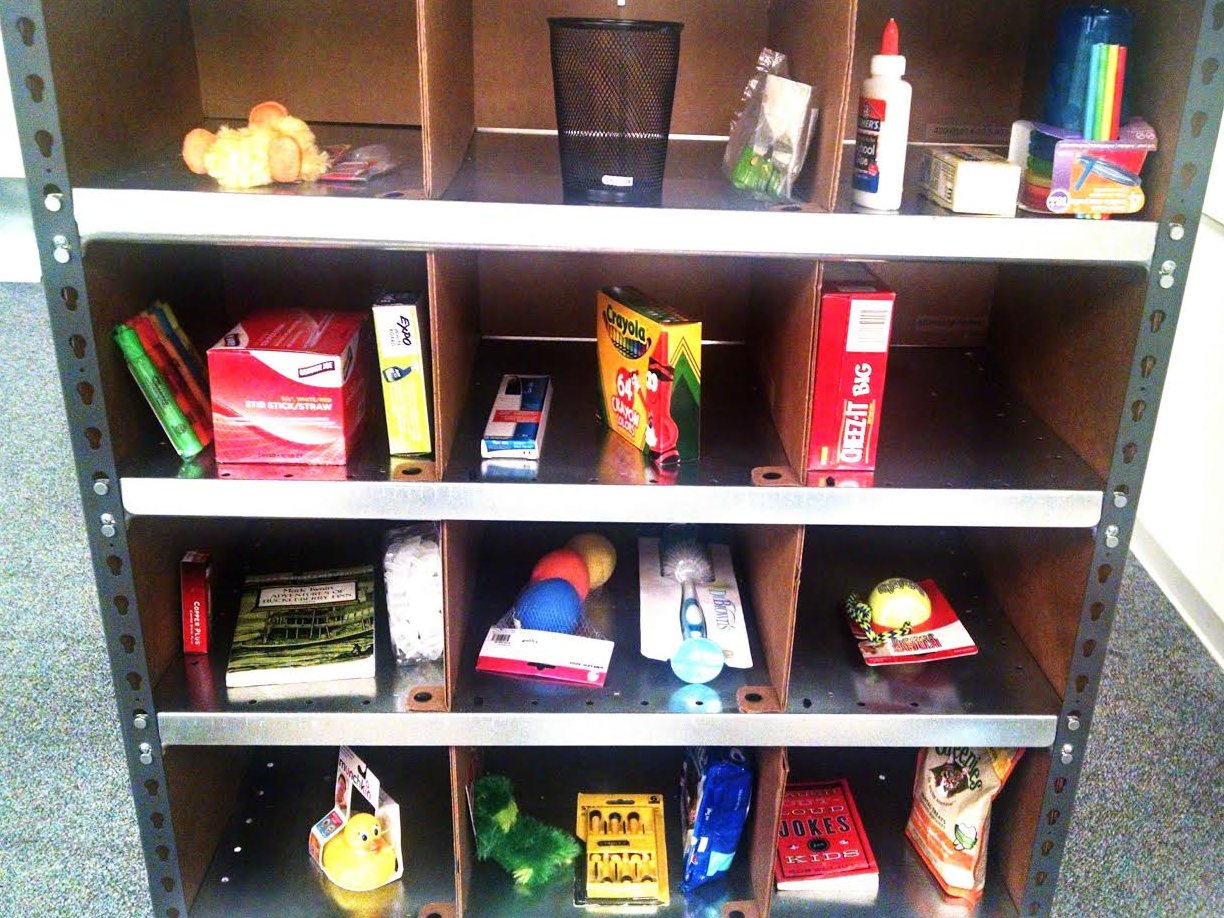} &
Rutgers APC RGB-D Dataset \cite{rennie-ral-2016} &
Kinect v1 &
24 &
10,368 &
Semi-manual ground truth alignment &
'16 \\
\bottomrule
\end{tabular}
\end{threeparttable}
\end{table*}

Some datasets \cite{shotton-cvpr-2013, meister-iros-2012, zhou-siggraph-2013, firman-cvpr-2016} use manually verified tracking from the Kinect itself as a ‘ground truth’ pose. This data is only suitable for tasks an order of magnitude harder than tracking, such as camera relocalization \cite{shotton-cvpr-2013} or voxel occupancy prediction \cite{firman-cvpr-2016}.

The difficulties involved with acquiring ground truth data can be circumvented with synthetic data.
The ICL-NUIM dataset \cite{handa-icra-2014} provides 8 camera trajectories for two synthetic indoor scenes, with camera paths taken from real hand-held camera trajectories.
While synthetic datasets may not be a perfect representation of our world, they allow users to more carefully control aspects such as motion blur and texture levels to gain introspection into their algorithm (see Section \ref{sec:synthetic} for further discussion).

\textbf{Scene reconstruction} is rarely evaluated directly, as good camera tracking usually corresponds to good reconstruction and camera paths are easier to obtain as ground truth than dense surfaces.
The synthetic ICL-NUIM dataset \cite{handa-icra-2014} is suitable for reconstruction evaluation, especially with additional camera paths provided by \cite{choi-cvpr-2015}.
More recently Wasenm\"uller \ea \cite{wasenmueller-wacv-2016} created a dataset containing ground truth camera motions and scene reconstructions from a laser scanner.
This is the only real-world dataset we are aware of with both these data, though the scenes are less diverse than \cite{sturm-iros-2012}.

Firman \ea \cite{firman-cvpr-2016} have a dataset of tabletop objects scanned so every visible surface is observed in the reconstruction.
This provides ground truth for the task of estimating the unobserved voxel occupancy from a depth image.


\subsection{Object pose estimation}
\begin{figure}[t]
\footnotesize
\begin{tabular}{m{2.5cm} m{4cm}}
\includegraphics[width=0.34\columnwidth]{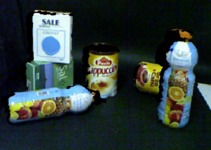}
&
\parbox{4cm}{\textbf{Realism:} \newmoon \fullmoon \fullmoon \vspace{4pt} \\
Laboratory scenarios, with a limited set of objects arranged by hand.\vspace{4pt} \\
Image from \protect{\cite{tombari-iros-2011}}}
\\
\includegraphics[width=0.34\columnwidth]{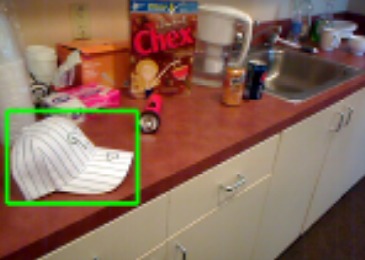}
&
\parbox{4cm}{\textbf{Realism:}  \newmoon \newmoon \fullmoon  \vspace{4pt} \\
Real-world scenes, but with furniture or objects artificially arranged.\vspace{4pt} \\
Image from \protect{\cite{lai-icra-2011}}
}
\\
\includegraphics[width=0.34\columnwidth]{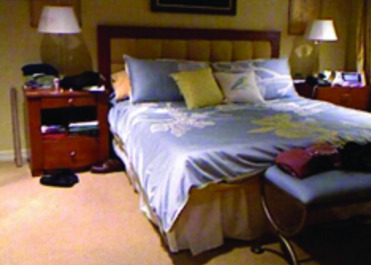}
&
\parbox{4cm}{\textbf{Realism:}  \newmoon \newmoon \newmoon  \vspace{4pt} \\
Real-world scenes with no interference by researchers. \vspace{4pt} \\
Image from \protect{\cite{silberman-eccv-2012}}}
\\
\end{tabular}
\caption{Semantic datasets described in Table \ref{tab:semantics} view the world in various levels of `realism', which we discretise into three categories.}
\vspace{-10pt}
\label{fig:realism}
\end{figure}

The problem of inferring the 6-DoF pose of an object is again a task which has been aided by the absolute scale provided by depth cameras.
Given \emph{a priori} a 3D model of an object, the aim is to find the transformation which best aligns it into the scene.
As with camera tracking it is hard to get ground truth for this type of challenge, which requires both a 3D model of the object and its pose in each image.
One solution has been to fix the target objects to a calibration board to allow for ground-truth tracking using the RGB channels \cite{hinterstoisser-accv-2012}, while \cite{salti-cviu-2014} and \cite{rennie-ral-2016} have the poses manually aligned.

These datasets, summarized in Table \ref{tab:objectpose}, feature tabletop-sized objects.
Acquiring 3D models, and ground truth poses, for larger objects is difficult, so works that have attempted this problem on a room scale typically find an alternative method of evaluation or rely on human annotations as an approximate ground truth \cite{song-cvpr-2015}.
Synthetic data could  be an avenue worth exploring here.


\subsection{Semantic labeling}
Semantic labeling of images and videos moves us to a more general understanding of the world.
Datasets with labels which could be used for semantic understanding are listed in Table \ref{tab:semantics}.
We give an indication of the `realism' of each dataset as a score out of three, explained in Figure \ref{fig:realism}.
Note that a low score here does not correspond to a worse or less useful dataset, as datasets with specially constructed scenarios can be vital for proving concepts, and they can often provide higher quality ground truth than fully natural scenes.

The 1449-frame subset of the NYUv2 dataset \cite{silberman-eccv-2012} with dense semantic labels has become a de-facto standard for indoor scene labeling.
The quality and variety of labels on this real-world dataset has helped make it one of the most highly used in the literature.
The SUN3D dataset \cite{xiao-iccv-2013} counters the single, static-frame modality of NYUv2 with object labels propagated through Kinect videos.
However, in spite of their effort, there are only 8 annotated sequences.

\begin{table*}[]
\centering
\footnotesize
\renewcommand\arraystretch{\stretchy}
\caption{Datasets for semantic reasoning and segmentation}
\label{tab:semantics}
\begin{threeparttable}
\begin{tabular}{m{0.9cm} >{\raggedright}m{4.4cm} m{2cm} c c m{5cm} c}
\toprule
&
& \textbf{Size}
& \textbf{Video?}
& \textbf{Realism\tnote{a}}
& \textbf{Labeling}
& \textbf{Year}
\\
\midrule
\includegraphics[height=\largeheight]{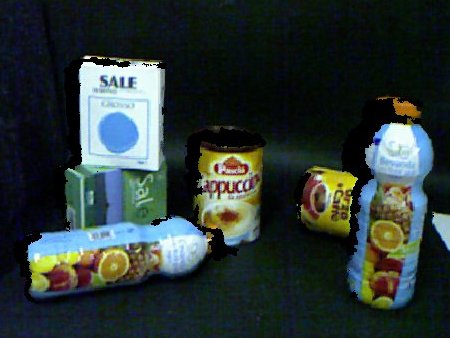}
 & 
RGB-D Semantic Segmentation Dataset \cite{tombari-iros-2011} & 
16 frames
 & 
 & 
\newmoon
\fullmoon
\fullmoon
 & 
Dense pixel labeling
 & 
'11
\\
\includegraphics[height=\largeheight]{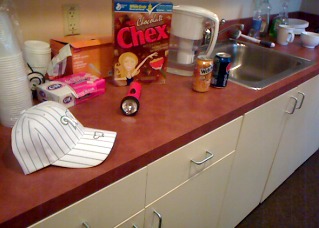}
 & 
RGBD Scenes dataset \cite{lai-icra-2011} & 
8 scenes
 & 
\checkmark
 & 
\newmoon
\newmoon
\fullmoon
 & 
Bounding box labeling of objects from the RGBD Objects dataset
 & 
'11
\\
\includegraphics[height=\largeheight]{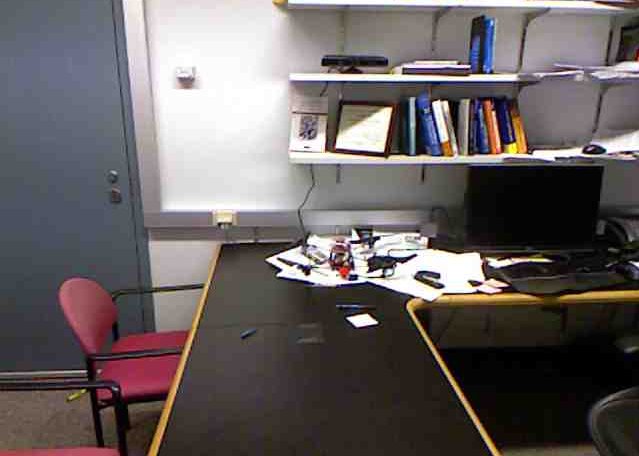}
 & 
Cornell-RGBD-Dataset \cite{koppula-nips-2011} & 
52 scenes
 & 
\checkmark
 & 
\newmoon
\newmoon
\newmoon
 & 
Semantic segmentation of reconstructed point cloud into 17 classes
 & 
'11
\\
\includegraphics[height=\largeheight]{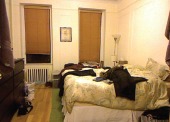}
 & 
NYUv1 \cite{silberman-iccv-2011} & 
2283 frames
 & 
-\tnote{b}
 & 
\newmoon
\newmoon
\newmoon
 & 
Dense pixel labeling
 & 
'11
\\
\includegraphics[height=\largeheight]{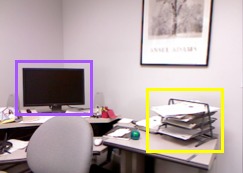}
 & 
Berkeley 3-D Object Dataset \cite{janoch-iccv-2011} & 
848 frames
 & 
 & 
\newmoon
\newmoon
\newmoon
 & 
Bounding box annotation
 & 
'11
\\
\includegraphics[height=\largeheight]{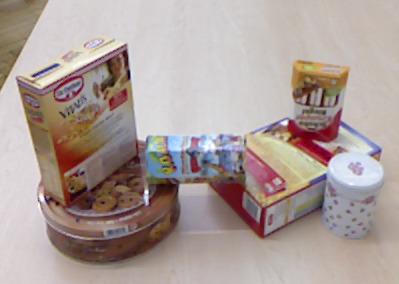}
 & 
Object segmentation dataset \cite{richtsfeld-iros-2012} & 
111 frames
 & 
 & 
\newmoon
\fullmoon
\fullmoon
 & 
Per-pixel segmentation into objects; no semantics
 & 
'12
\\
\includegraphics[height=\largeheight]{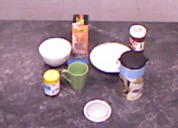}
 & 
MPII Multi-Kinect dataset \cite{wandi-eccv-2012} & 
2240 frames total from 4 Kinects
 & 
 & 
\newmoon
\fullmoon
\fullmoon
 & 
Polygon segmentation of objects arranged on kitchen worktop
 & 
'12
\\
\includegraphics[height=\largeheight]{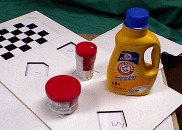}
 & 
Willow garage dataset \cite{aldoma-www-2012} & 
\texttildelow160 frames
 & 
 & 
\newmoon
\fullmoon
\fullmoon
 & 
Dense pixel labeling
 & 
'12
\\
\includegraphics[height=\largeheight]{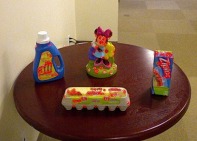}
 & 
Object Disappearance for Object Discovery \cite{mason-iros-2012} & 
1231 frames
 & 
\checkmark
 & 
\newmoon
\newmoon
\fullmoon
 & 
Ground truth object segmentations of objects of interest
 & 
'12
\\
\includegraphics[height=\largeheight]{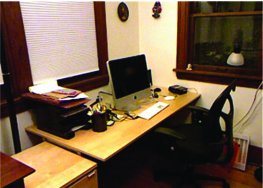}
 & 
NYUv2 \cite{silberman-eccv-2012} & 
1449 frames from 464 scenes
 & 
-\tnote{b}
 & 
\newmoon
\newmoon
\newmoon
 & 
Dense pixel labeling. A synthetic re-creation of the 3D scenes also exists \cite{guo-iccv-2013}
 & 
'12
\\
\includegraphics[height=\largeheight]{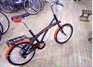}
 & 
RGBD Dataset for Category Modeling \cite{zhang-cvpr-2013} & 
900 frames
 & 
 & 
\newmoon
\newmoon
\fullmoon
 & 
Which of 7 categories the dominant object in each image is in
 & 
'13
\\
\includegraphics[height=\largeheight]{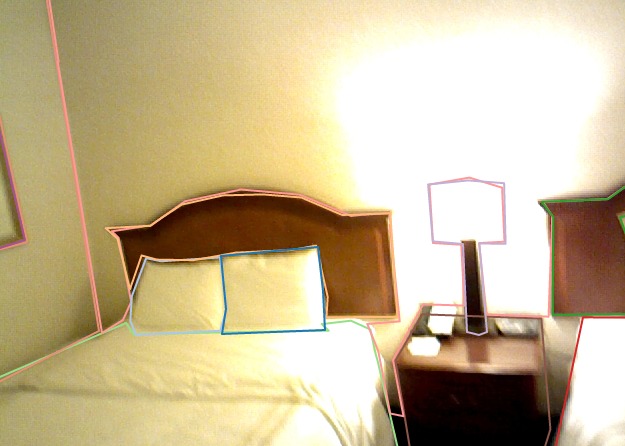}
 & 
SUN3D \cite{xiao-iccv-2013} & 
8 scenes
 & 
\checkmark
 & 
\newmoon
\newmoon
\newmoon
 & 
Polygon labels. 8 scenes labeled, though full dataset has more
 & 
'13
\\
\includegraphics[height=\largeheight]{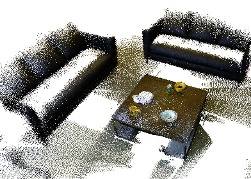}
 & 
RGBD Scenes dataset v2 \cite{lai-icra-2014} & 
14 scenes
 & 
\checkmark
 & 
\newmoon
\newmoon
\fullmoon
 & 
Items from the RGBD Objects dataset labeled on reconstructed point cloud
 & 
'14
\\
\includegraphics[height=\largeheight]{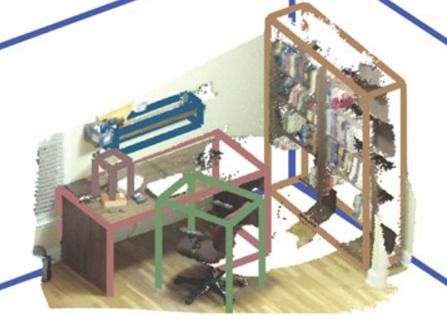}
 & 
SUN RGB-D \cite{song-cvpr-2015} & 
10,335 frames\tnote{c}
 & 
 & 
\newmoon
\newmoon
\newmoon
 & 
3D object bounding boxes, and polygons on 2D images
 & 
'15
\\
\includegraphics[height=\largeheight]{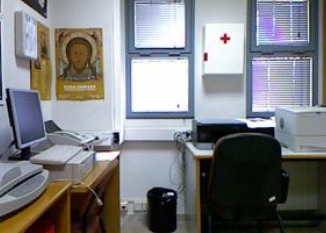}
 & 
ViDRILO \cite{martinezgomez-ijrr-2015} & 
22454 frames from 5 scenes
 & 
\checkmark
 & 
\newmoon
\newmoon
\newmoon
 & 
Semantic category of frame, plus which objects are visible in each frame
 & 
'15
\\
\includegraphics[height=\largeheight]{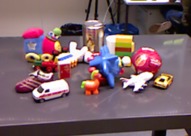}
 & 
Toy dataset \cite{ikkala-visapp-2016} & 
449 frames
 & 
 & 
\newmoon
\fullmoon
\fullmoon
 & 
Per-pixel segmentation into objects; no semantics
 & 
'16
\\

\bottomrule
\end{tabular}
\begin{tablenotes}
  \item[a] See Figure \ref{fig:realism}
  \item[b] Extended version of dataset has video, but labels are only present in subset described here.
  \item[c] Combines new Kinect v2 frames with new labels on existing datasets \cite{silberman-eccv-2012, janoch-iccv-2011, xiao-iccv-2013}
\end{tablenotes}
\end{threeparttable}
\end{table*}

We note that all these semantic datasets, even those with videos, depict a static world.
This contrasts with our \emph{dynamic} world, an area which is explored by datasets designed for tracking.

\begin{table*}[]
\renewcommand\arraystretch{\stretchy}
 \newcommand{\rot}[1]{\multicolumn{1}{c}{\adjustbox{angle=30,lap=\width-1em}{#1}}‌​}

\centering
\footnotesize
\caption{Datasets representing activities and gestures}
\label{tab:activites}
\begin{threeparttable}
\begin{tabular}{m{0.7cm} m{4.5cm} c c c c m{6cm} m{0.5cm}}
\toprule
&
& \rot{\textbf{\# Subjects}}
& \rot{\textbf{\# Actions}}
& \rot{\textbf{\# Videos}}
& \rot{\textbf{Skeleton\tnote{a}}}
& \textbf{Examples of actions}
& \textbf{Year}
\\
\midrule
\includegraphics[height=\smallheight]{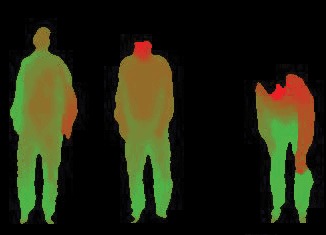}
 & 
MSR Action3D \cite{li-cvprw-2010} & 
10
 & 
20
 & 
567
 & 
\checkmark
 & 
\eg \emph{high arm wave, side kick, jogging}
 & 
'10
\\
\includegraphics[height=\smallheight]{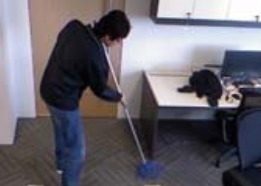}
 & 
RGBD-HuDaAct \cite{ni-iccvw-2011} & 
30
 & 
12
 & 
1189
 & 
 & 
\eg \emph{get up, enter room, stand up, mop the floor}
 & 
'11
\\
\includegraphics[height=\smallheight]{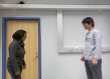}
 & 
SBU Kinect Interaction Dataset \cite{yun-cvprw-2012} & 
7
 & 
8
 & 
300
 & 
\checkmark
 & 
Two people interacting \eg \emph{approaching, departing}
 & 
'12
\\
\includegraphics[height=\smallheight]{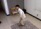}
 & 
ACT4\textsuperscript{2} \cite{cheng-eccvw-2012} & 
24
 & 
14
 & 
6844
 & 
 & 
4 Kinects filming. Actions: \eg \emph{collapse, reading}
 & 
'12
\\
\includegraphics[height=\smallheight]{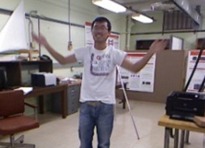}
 & 
UTKinect-Action \cite{xia-cvprw-2012} & 
10
 & 
10
 & 
200
 & 
\checkmark
 & 
\eg \emph{walk, sit down, stand up, carry, clap hands}
 & 
'12
\\
\includegraphics[height=\smallheight]{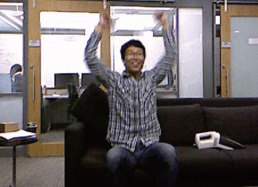}
 & 
MSRDailyActivity3D \cite{wang-cvpr-2012} & 
10
 & 
16
 & 
320
 & 
\checkmark
 & 
\eg \emph{drink, eat, read book}
 & 
'12
\\
\includegraphics[height=\smallheight]{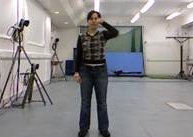}
 & 
G3D Gaming Action Dataset \cite{bloom-cvprw-2012} & 
10
 & 
20
 & 
600
 & 
\checkmark
 & 
Typical gaming actions
 & 
'12
\\
\includegraphics[height=\smallheight]{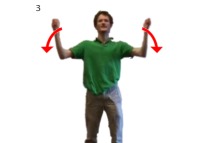}
 & 
MSRC-12 Kinect gesture \cite{fothergill-chi-2012} & 
30
 & 
12
 & 
594
 & 
\checkmark
 & 
Arm gestures
 & 
'12
\\
\includegraphics[height=\smallheight]{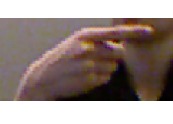}
 & 
MSRGesture3D \cite{kurakin-eusipco-2012} & 
10
 & 
12
 & 
336
 & 
 & 
American Sign Language
 & 
'12
\\
\includegraphics[height=\smallheight]{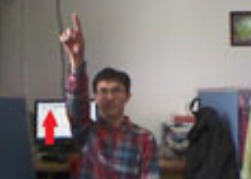}
 & 
ChaLearn Gesture Challenge \cite{guyon-adiaa-2012} & 
20
 & 
~850
 & 
50000
 & 
 & 
Many, \eg \emph{diving signals and mudras}
 & 
'12
\\
\includegraphics[height=\smallheight]{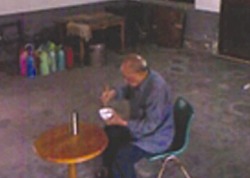}
 & 
Senior Activity Recognition (RGBD-SAR) \cite{yang-cc-2013} & 
30
 & 
9
 & 
810
 & 
\checkmark
 & 
Older people performing activities \eg \emph{sit down, eat, walk, stand up}
 & 
'13
\\
\includegraphics[height=\smallheight]{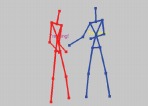}
 & 
K3HI \cite{hu-mpe-2013} & 
15
 & 
8
 & 
320
 & 
\checkmark
 & 
Two humans interacting \eg \emph{approaching, punching}
 & 
'13
\\
\includegraphics[height=\smallheight]{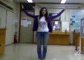}
 & 
UPCV action dataset \cite{theodorakopoulos-jvcir-2013} & 
20
 & 
10
 & 
400
 & 
\checkmark
 & 
\eg \emph{walk, wave, scratch head, phone, cross arms}
 & 
'13
\\
\includegraphics[height=\smallheight]{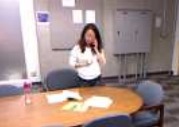}
 & 
DML-SmartAction \cite{amiri-hc-2013} & 
16
 & 
12
 & 
932
 & 
 & 
Continuous recording. \eg \emph{writing, sit down, walk, clean table, stand up}
 & 
'13
\\
\includegraphics[height=\smallheight]{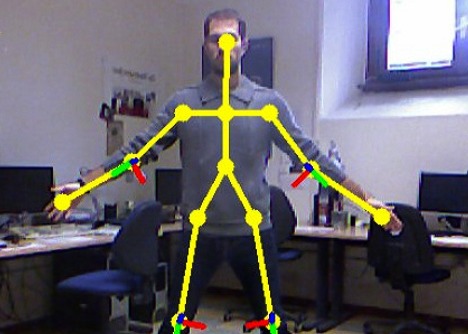}
 & 
Florence 3D actions dataset \cite{seidenari-cvprw-2013} & 
10
 & 
9
 & 
215
 & 
\checkmark
 & 
\eg \emph{wave, drinking, answer phone, clap, stand up}
 & 
'13
\\
\includegraphics[height=\smallheight]{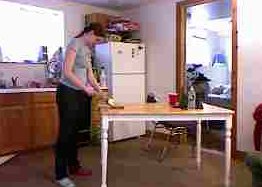}
 & 
Cornell activty 60/120 \cite{sung-pair-2011, koppula-ijrr-2013} & 
4
 & 
12/10
 & 
60/120
 & 
\checkmark
 & 
\eg \emph{brushing teeth, drinking, talking on couch}
 & 
'13
\\
\includegraphics[height=\smallheight]{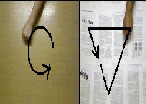}
 & 
Sheffield KInect Gesture (SKIG) \cite{liu-ijcai-2013} & 
6
 & 
10
 & 
1080
 & 
 & 
Hand gestures \eg \emph{circle, up-down, comehere}
 & 
'13
\\
\includegraphics[height=\smallheight]{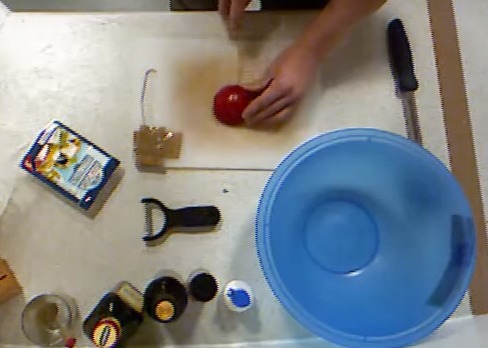}
 & 
50 Salads \cite{stein-ubicomp-2013} & 
25
 & 
2
 & 
50
 & 
 & 
Each person prepares two salads. Accelerometer on utensils
 & 
'13
\\
\includegraphics[height=\smallheight]{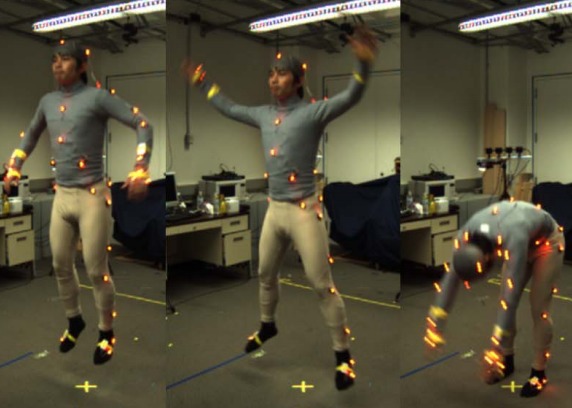}
 & 
Berkeley Multimodal Human Action \cite{ofli-wacv-2013} & 
12
 & 
11
 & 
660
 & 
\checkmark\checkmark
 & 
\eg \emph{jumping, bending, punching}
 & 
'13
\\
\includegraphics[height=\smallheight]{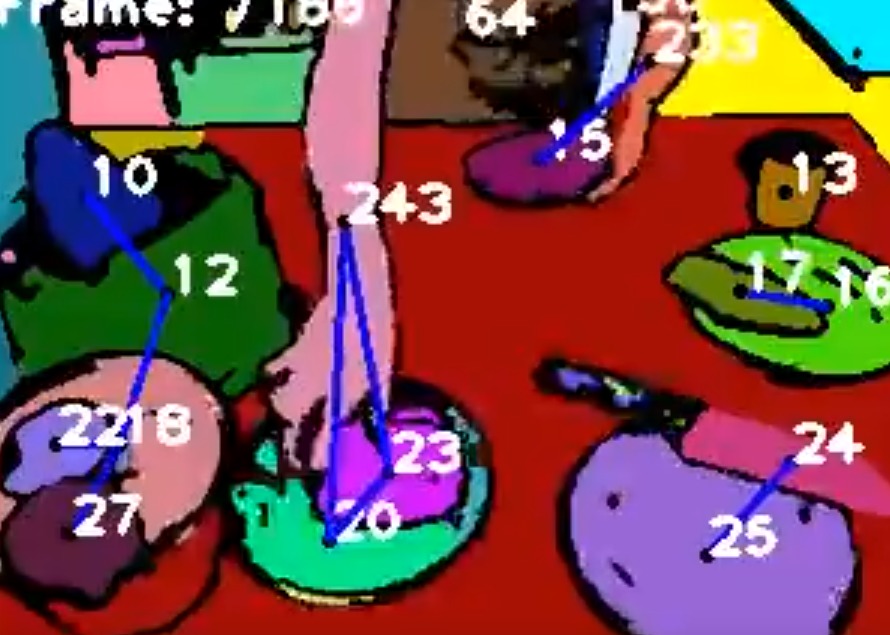}
 & 
Manipulation Action Dataset \cite{aksor-ras-2014} & 
5
 & 
28
 & 
140
 & 
 & 
Manipulation actions \eg \emph{cutting}, plus sequences of actions. Semantic segmentation of frames.
 & 
'14
\\
\includegraphics[height=\smallheight]{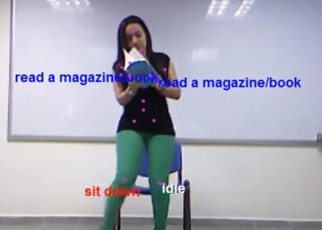}
 & 
Composable activities dataset \cite{lillo-cvpr-2014} & 
14
 & 
16
 & 
693
 & 
\checkmark
 & 
\eg \emph{throw, talk on phone, walk, wave, crouch, punch}
 & 
'14
\\
\includegraphics[height=\smallheight]{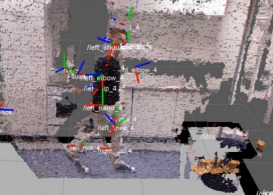}
 & 
TUM Morning Routine Dataset \cite{karg-aamas-2014} & 
1
 & 
-
 & 
-\tnote{b}
 & 
\checkmark
 & 
Typical morning routine activities
 & 
'14
\\
\includegraphics[height=\smallheight]{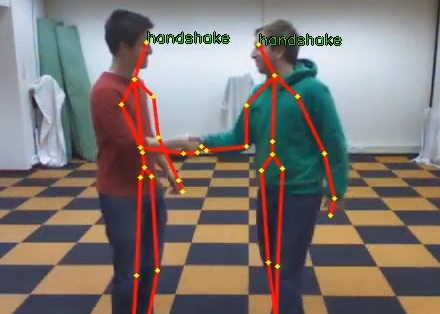}
 & 
ShakeFive \cite{gemeren-eccv-2014} & 
37
 & 
2
 & 
100
 & 
\checkmark
 & 
Hand shake or high-five between two individuals
 & 
'14
\\
\includegraphics[height=\smallheight]{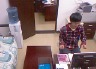}
 & 
Office activity dataset \cite{wang-icm-2014} & 
$>$10
 & 
20
 & 
1180
 & 
 & 
\eg \emph{mopping, sleeping, finding-objects, chatting}
 & 
'14
\\
\includegraphics[height=\smallheight]{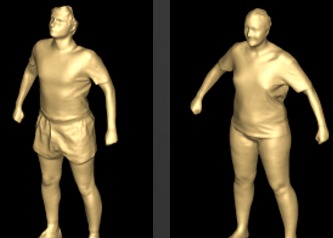}
 & 
Human3.6M \cite{ionescu-pami-2014} & 
11
 & 
17
 & 
-\tnote{b}
 & 
\checkmark\checkmark
 & 
\eg \emph{Discussion, smoking, taking photo}
 & 
'14
\\
\includegraphics[height=\smallheight]{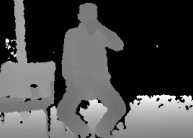}
 & 
MSR 3D Online Action \cite{yu-accv-2014} & 
24
 & 
7
 & 
-\tnote{b}
 & 
 & 
\eg \emph{drinking, eating, using laptop}
 & 
'14
\\
\includegraphics[height=\smallheight]{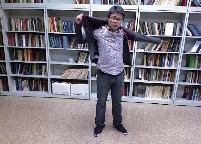}
 & 
Northwestern-UCLA Multiview Action 3D \cite{wang-cvpr-2014} & 
10
 & 
10
 & 
-\tnote{b}
 & 
\checkmark
 & 
Three Kinects filming. Actions: \eg \emph{stand up, throw}
 & 
'14
\\
\includegraphics[height=\smallheight]{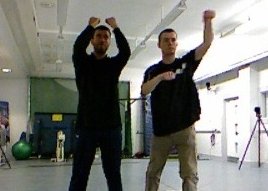}
 & 
G3Di Gaming Interaction Dataset \cite{bloom-eccvw-2014} & 
12
 & 
17
 & 
-\tnote{b}
 & 
\checkmark
 & 
Humans interacting with computer game
 & 
'14
\\
\includegraphics[height=\smallheight]{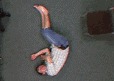}
 & 
UR Fall Detection \cite{kwolek-cmpb-2014} & 
?
 & 
1
 & 
70
 & 
 & 
Humans falling over. Two Kinects. Accelerometer from human
 & 
'14
\\
\includegraphics[height=\smallheight]{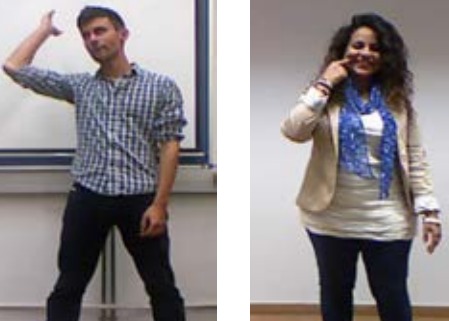}
 & 
Montalbano Gesture \cite{escalera-eccv-2014} & 
27
 & 
20
 & 
13858
 & 
\checkmark
 & 
Italian hand gestures
 & 
'14
\\
\includegraphics[height=\smallheight]{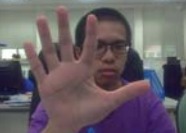}
 & 
LaRED Hand Gesture Dataset \cite{hsiao-msc-2014} & 
10
 & 
27
 & 
810
 & 
 & 
Modified American Sign Language
 & 
'14
\\
\includegraphics[height=\smallheight]{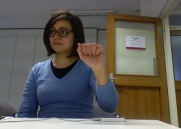}
 & 
LTTM MS Kinect and Leap Motion \cite{marin-icip-2014} & 
14
 & 
10
 & 
1400
 & 
 & 
American Sign Language, recorded using Kinect and the Leap Motion
 & 
'14
\\
\includegraphics[height=\smallheight]{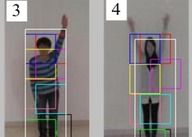}
 & 
TJU dataset \cite{liu-sp-2015} & 
22
 & 
22
 & 
1936
 & 
\checkmark
 & 
\eg \emph{boxing, one hand wave, forward bend, sit down}
 & 
'15
\\

\vspace{6pt} \\
\multicolumn{7}{c}{{\normalsize \texttt{Continued overleaf $\downarrow$}}} \\
\vspace{6pt} \\
\bottomrule
\end{tabular}
\begin{tablenotes}
  \item[a] \checkmark = 2D skeleton joint positions labeled on video frames; \checkmark\checkmark = 3D skeleton joint positions acquired from MoCap system
  \item[b] These datasets feature continuous footage, so the discrete number of videos is less meaningful here.
\end{tablenotes}
\end{threeparttable}
\end{table*}


\begin{table*}[]
\renewcommand\arraystretch{\stretchy}
 \newcommand{\rot}[1]{\multicolumn{1}{c}{\adjustbox{angle=45,lap=\width-1em}{#1}}‌​}

\centering
\footnotesize
\begin{threeparttable}
\begin{tabular}{m{0.7cm} m{4.5cm} c c c c m{5.5cm} m{0.5cm}}
\toprule
\vspace{6pt} \\
\multicolumn{7}{c}{{\normalsize $\hookrightarrow$ \texttt{Continued from previous page}}} \\
\vspace{6pt}
\includegraphics[height=\smallheight]{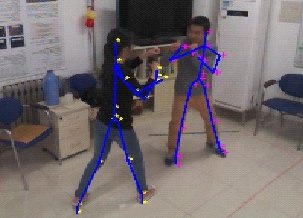}
 & 
M\textsuperscript{2}I dataset \cite{xu-icm-2015} & 
22
 & 
22
 & 
1760
 & 
\checkmark
 & 
Two people interacting, \eg \emph{walk together}
 & 
'15
\\
\includegraphics[height=\smallheight]{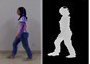}
 & 
Multi-view TJU \cite{liu-sp-2015} & 
20
 & 
22
 & 
7040
 & 
\checkmark
 & 
Front and side view Kinects. Actions as TJU dataset
 & 
'15
\\
\includegraphics[height=\smallheight]{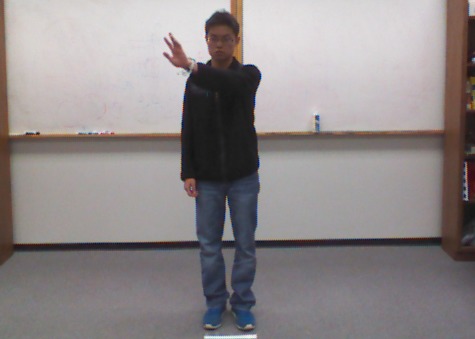}
 & 
UTD Multimodal Human Action \cite{chen-icip-2015} & 
8
 & 
27
 & 
861
 & 
\checkmark
 & 
Accelerometer data. Actions: \eg \emph{wave, boxing}
 & 
'15
\\
\includegraphics[height=\smallheight]{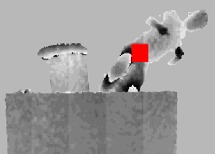}
 & 
TST Fall Detection ver. 1/ver. 2 \cite{gasparrini-sensors-2014, gasparrini-ictinv-2015} & 
4/11
 & 
2
 & 
20/111
 & 
\checkmark
 & 
Humans falling over
 & 
'15
\\
\includegraphics[height=\smallheight]{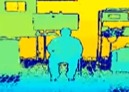}
 & 
TST TUG \cite{cippitelli-icc-2015} & 
20
 & 
?
 & 
60
 & 
\checkmark
 & 
Timed Up and Go tests
 & 
'15
\\
\includegraphics[height=\smallheight]{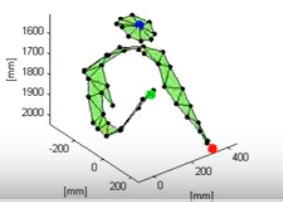}
 & 
TST Intake Monitoring ver 1/ver 2 \cite{gasparrini-techall-2015} & 
35
 & 
?
 & 
35/60
 & 
 & 
Humans simulating eating
 & 
'15
\\
\includegraphics[height=\smallheight]{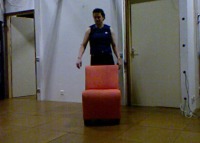}
 & 
Life activities with occlusions \cite{dib-iros-2015} & 
1
 & 
-
 & 
12
 & 
\checkmark\checkmark
 & 
No specific actions
 & 
'15
\\
\includegraphics[height=\smallheight]{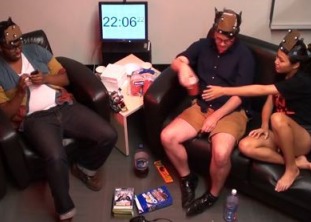}
 & 
Background activity dataset \cite{freeman-arxiv-2015} & 
52
 & 
4
 & 
-\tnote{b}
 & 
\checkmark\checkmark
 & 
Humans natually interacting in semi-natural environment
 & 
'15
\\
\includegraphics[height=\smallheight]{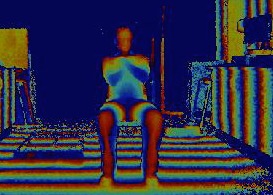}
 & 
K3Da \cite{leightley-apsipa-2015} & 
53
 & 
13
 & 
?
 & 
\checkmark
 & 
To assess human health, \eg \emph{leg jump, walking}
 & 
'15
\\
\includegraphics[height=\smallheight]{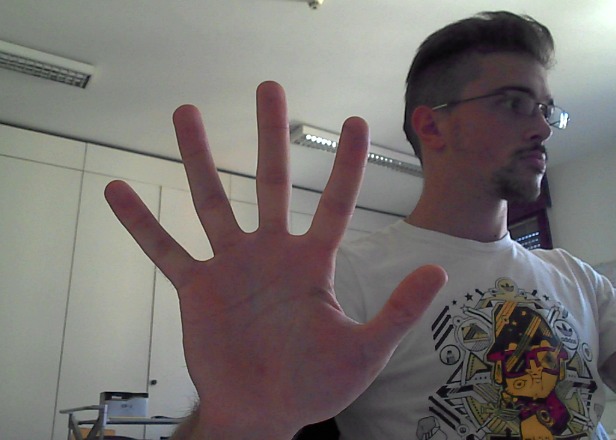}
 & 
LTTM Creative Senz3D \cite{memo-stag-2015} & 
4
 & 
11
 & 
1320
 & 
 & 
Hand gestures \eg \emph{`OK'}
 & 
'15
\\
\includegraphics[height=\smallheight]{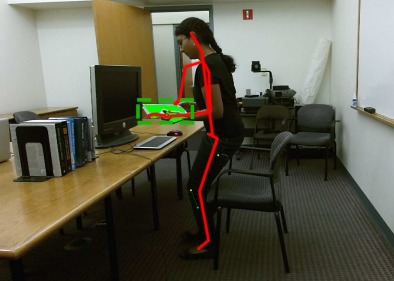}
 & 
Watch-n-Patch \cite{wu-cvpr-2015} & 
7
 & 
21
 & 
458
 & 
 & 
A sequence of actions \eg \emph{making drink}
 & 
'15
\\
\includegraphics[height=\smallheight]{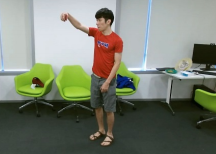}
 & 
NTU RGB+D \cite{shahroudy-cvpr-2016} & 
40
 & 
60
 & 
56,000
 & 
\checkmark
 & 
\eg \emph{drinking, eating, sneezing, staggering, punching, kicking}
 & 
'16
\\

\bottomrule
\end{tabular}
\vspace{-10pt}
\end{threeparttable}
\end{table*}


\begin{table*}[!b]
\renewcommand\arraystretch{0.5}
\centering
\footnotesize
\caption{Datasets of faces for pose and recognition}
\label{tab:faces}
\begin{threeparttable}
\begin{tabular}{m{0.9cm} >{\raggedright}m{2.6cm} m{0.8cm} >{\raggedright}m{1.8cm} >{\raggedright}m{3.8cm} >{\raggedright}m{3.8cm} m{0.5cm}}
\toprule
&
& \textbf{Subjects}
& \textbf{Sensor}
& \textbf{Description}
& \textbf{Labeling}
& \textbf{Year}
\\
\midrule
\includegraphics[height=\largeheight]{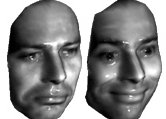}
 & 
Human Face \cite{bronstein-vcg-2007}
 & 
1
 & 
Structured light scanner
 & 
15 expressions performed by one face
 & 
-
 & 
'07
\\
\includegraphics[height=\largeheight]{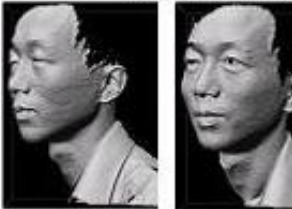}
 & 
CASIA 3D Face Database \cite{casia-minolta-2008}
 & 
123
 & 
Minolta Vivid 910
 & 
4624 images of various expressions, poses and lighting
 & 
Expression performed
 & 
'08
\\
\includegraphics[height=\largeheight]{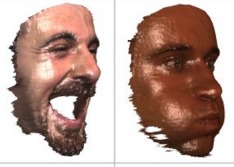}
 & 
Bosphorus Database \cite{savran-cost-2008}
 & 
105
 & 
Inspeck Mega Capturor II 3D
 & 
Faces performing expressions at different rotations
 & 
Expression and pose
 & 
'08
\\
\includegraphics[height=\largeheight]{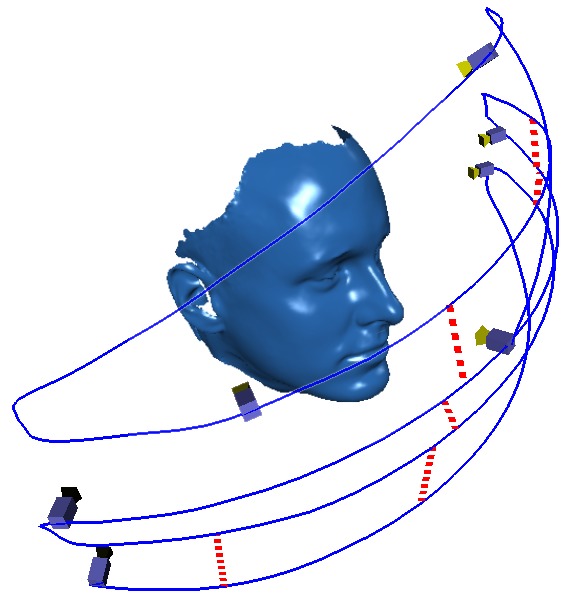}
 & 
ETH Face Pose Range Image Data Set \cite{breitenstein-cvpr-2008}
 & 
20
 & 
Custom active stereo setup
 & 
Videos of face in various poses
 & 
Nose position and coordinate frame at the nose
 & 
'08
\\
\includegraphics[height=\largeheight]{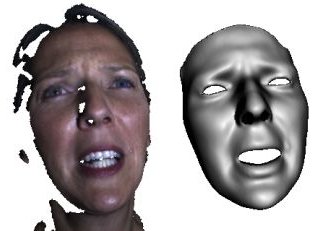}
 & 
B3D(AC)\^{}2 \cite{fanelli-tom-2010}
 & 
14
 & 
Custom active stereo setup
 & 
Recordings of humans speaking
 & 
Perceived emotions. Audio labeled with phonemes
 & 
'10
\\
\includegraphics[height=\largeheight]{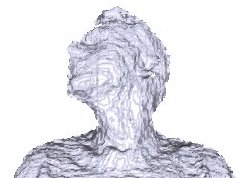}
 & 
Biwi Kinect Head Pose Database \cite{fanelli-dadm-2011}
 & 
20
 & 
Kinect v1
 & 
People moving their heads in different directions
 & 
3D position of the head and its rotation
 & 
'11
\\
\includegraphics[height=\largeheight]{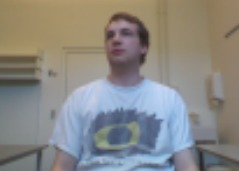}
 & 
VAP RGB-D Face Database \cite{hog-sitibs-2012}
 & 
31
 & 
Kinect v1
 & 
1581 images of people doing different poses in front a camera
 & 
Which person is in shot, and a discretised gaze direction
 & 
'12
\\
\includegraphics[height=\largeheight]{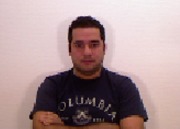}
 & 
3D Mask Attack Dataset \cite{erdogmus-btas-2013}
 & 
17
 & 
Kinect v1
 & 
Some frames are of person with a face mask of someone else
 & 
Person's identity, and if `spoofing' is occurring. Eye positions
 & 
'13
\\
\includegraphics[height=\largeheight]{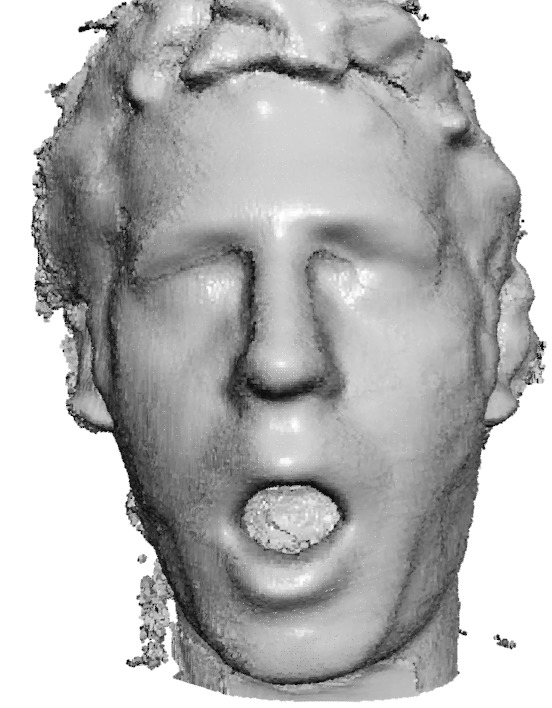}
 & 
Face Warehouse \cite{chen-vcg-2014}
 & 
150
 & 
Kinect v1
 & 
People performing expressions
 & 
Which of 20 expressions, plus 74 landmarks and meshes
 & 
'14
\\
\includegraphics[height=\largeheight]{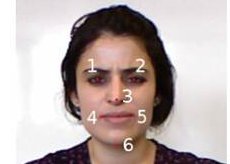}
 & 
Eurecom Kinect Face Dataset \cite{min-smc-2014}
 & 
52
 & 
Kinect v1
 & 
Faces with different expressions, occlusion and illumination
 & 
Expression type, and six facial landmark locations
 & 
'14
\\
\includegraphics[height=\largeheight]{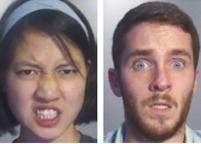}
 & 
VT-KFER \cite{aly-icb-2015}
 & 
32
 & 
Kinect v1
 & 
7 facial expressions labeled, in scripted and unscripted scenarios
 & 
Percieved expression
 & 
'15
\\

\bottomrule
\end{tabular}
\end{threeparttable}
\end{table*}

\subsection{Tracking}
Tracking datasets feature videos of \emph{dynamic} worlds, where the aim is to detect where an object is in each frame.
We know of only four datasets explicitly designed for this purpose, all of which use bounding boxes as annotations.
The Princeton Tracking Benchmark \cite{song-iccv-2013} contains 100 videos of moving objects, such as dogs and toy cars.
The RGB-D people dataset \cite{spinello-iros-2011, luber-iros-2011}, the Kinect Tracking Precision dataset \cite{munaro-iros-2012} and the RGBD Pedestrian Dataset \cite{bagautdinov-cvpr-2015} all track humans.

Other datasets contain labels appropriate for tracking: two semantic scene datasets \cite{xiao-iccv-2013, lai-icra-2011} have static objects labeled through video as the camera moves, while the 6-DOF object pose annotations in \cite{hinterstoisser-accv-2012} could also be useful.

\subsection{Activities and gestures}
\label{sec:actions}

Given the original use case of the Kinect as a sensor designed for human interaction, it is inevitable that much research would focus on recognizing gestures and activities from videos.
See Table \ref{tab:activites} for an overview of the large number of datasets in this area, and we refer the interested reader to \cite{zhang-arxiv-2016} for a more detailed survey of this field.

Actions being performed include sign language \cite{kurakin-eusipco-2012}, Italian hand gestures \cite{escalera-eccv-2014} and common daily actions such as standing up, drinking and reading \cite{sung-pair-2011, koppula-ijrr-2013, ni-iccvw-2011, wang-cvpr-2012, li-cvprw-2010, chen-icip-2015}.
Three datasets of humans falling over \cite{kwolek-cmpb-2014, gasparrini-sensors-2014, gasparrini-ictinv-2015} reflect an interest in use of RGBD sensors for monitoring vulnerable humans in their daily lives.
Others are more niche: 50 Salads \cite{stein-ubicomp-2013} features over 4 hours of people preparing mixed salads.
Four datasets stand out for capturing humans with a full MoCap setup \cite{dib-iros-2015, freeman-arxiv-2015, ofli-wacv-2013, ionescu-pami-2014}, while the Manipulation Action Dataset \cite{aksor-ras-2014} is unique in providing semantic segmentation of objects as they are manipulated.
By far the largest gesture and action datasets are the ChaLearn gesture challenge \cite{guyon-adiaa-2012} and NTU RGB+D \cite{shahroudy-cvpr-2016}, each with around 50,000 videos.



Many of these datasets suffer from being filmed in the confines of an office or laboratory, with researchers performing the actions.
Filming real people at work and home would help prevent dataset bias and provide a more believable baseline for activity and gesture recognition.

\subsection{Faces}
Early face datasets focused on the method of acquisition (\eg \cite{zhang-siggraph-2004}) or tended to be quite small (\eg \cite{bronstein-vcg-2007}).
The field has now expanded to include datasets for identity recognition \cite{erdogmus-btas-2013}, pose regression \cite{breitenstein-cvpr-2008, fanelli-dadm-2011}, and those where the expressions or emotions are to be inferred~\cite{fanelli-tom-2010, min-smc-2014}.
We summarize these in Table~\ref{tab:faces}, and more details on some of these datasets can be found in~\cite{aly-icb-2015}.
As front-facing depth cameras become installed in laptops and tablets we expect this area of research to continue to gain attention.

\subsection{Recognition}
Like datasets of actions, datasets designed for human recognition (Table \ref{tab:id}) typically film people performing activities such as walking.
However, the aim now is to recognize the identity, gender or other attributes about the subjects, rather than the activity they are performing.

\section{Future areas for datasets}
\label{sec:future}

In Section \ref{sec:present} we reviewed the past and the present of RGBD datasets.
We now look to the future, and identify underexplored `gaps in the market'.

\subsection{Synthetic data}
\label{sec:synthetic}
Aside from a few examples such as ICL-NUIM \cite{handa-icra-2014} and SceneNet \cite{handa-arxiv-2015}, synthetic data has received relatively little attention  for vision problems with depth cameras.
Yet such artificial data can offer many advantages. Ground truth for tasks such as segmentation, reconstruction, tracking and camera or object pose is perfect and available with no requirement for expensive human labeling.
Sequences can be recaptured with carefully adjusted parameters, \eg motion blur and lighting changes, for algorithm introspection. It is also possible to create scenarios difficult to capture in real life, for example car crashes.

While sensor noise can be emulated \cite{handa-icra-2014, gschwandtner-isvc-2011, firman-iros-2011, meister-vmv-2013}, it can be very difficult for synthetic scenes to capture the true properties of the real world.
One option is to use existing 3D assets.
The synthetic Sintel dataset \cite{butler-eccv-2012}, for example, has been used for RGB tasks such as optical flow.
With depth channels now available this may yet find a use in the RGBD community.
Another route is to use generative models of scenes, following work on scene synthesis  \cite{fisher-siggraph-2012, handa-arxiv-2015}.



\begin{table*}[]
\renewcommand\arraystretch{\stretchy}
\centering
\footnotesize
\caption{Datasets for human recognition}
\label{tab:id}
\begin{threeparttable}
\begin{tabular}{m{1cm} m{3cm} c m{5cm} m{4cm} m{1cm}}
\toprule
&
& \textbf{Subjects}
& \textbf{Description}
& \textbf{Labeling}
& \textbf{Year}
\\
\midrule
\includegraphics[height=\largeheight]{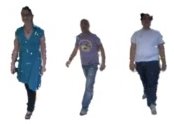}
 & 
RGB-D Person Re-identification \cite{barbosa-eccvw-2012}
 & 
79
 & 
Humans walking, where subjects change clothes between sessions
 & 
2D skeleton positions. Which human is in each video
 & 
'12
\\
\includegraphics[height=\largeheight]{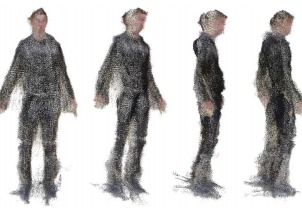}
 & 
IAS-Lab RGBD-ID Dataset \cite{munaro-icra-2014}
 & 
11
 & 
Humans walking, where subjects change clothes (or room) between sessions
 & 
2D skeleton positions. Which human is in each video
 & 
'14
\\
\includegraphics[height=\largeheight]{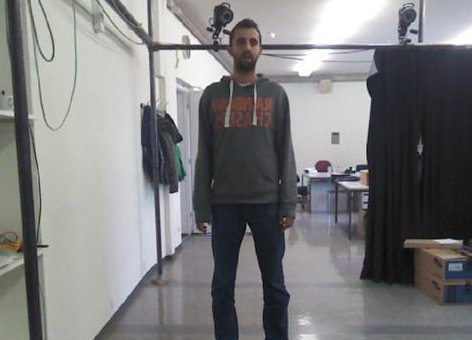}
 & 
BIWI RGBD-ID Dataset \cite{munaro-icra-2014}
 & 
50
 & 
Humans moving, where subjects change clothes (or room) between sessions
 & 
2D skeleton positions. Which human is in each video
 & 
'14
\\
\includegraphics[height=\largeheight]{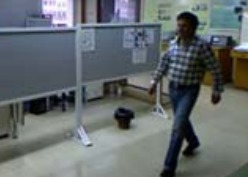}
 & 
UPCV Gait dataset \cite{kastaniotis-prl-2015}
 & 
30
 & 
Each human walks down corridor multiple times
 & 
Identity and gender of each person
 & 
'15
\\

\bottomrule
\end{tabular}
\end{threeparttable}
\end{table*}

\subsection{Full voxel occupancy}
\begin{figure}[t]
  \footnotesize
  \setlength\extrarowheight{38pt}
  \vspace{-20pt}
  \begin{tabular}{m{4cm} m{3cm}}
  \includegraphics[width=0.49\columnwidth]{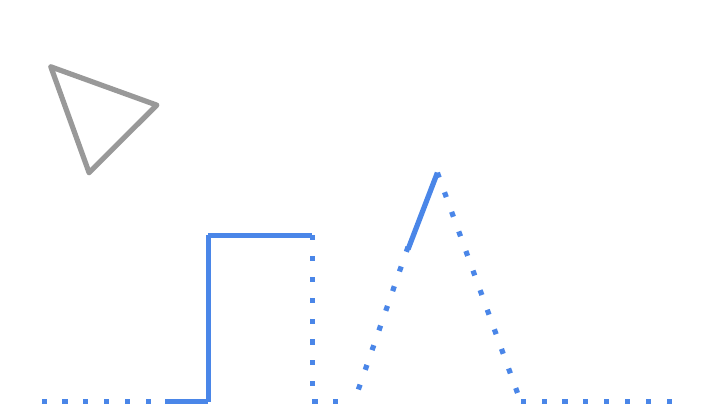}
  &
  \parbox{4cm}{
    (a) Early RGBD datasets focussed on single images of scenes, representing them in 2\nicefrac{1}{2}D. \vspace{4pt} \\
    Examples: \cite{silberman-eccv-2012, janoch-iccv-2011, richtsfeld-iros-2012} }
  \\
  \vspace{8pt}

  \includegraphics[width=0.49\columnwidth]{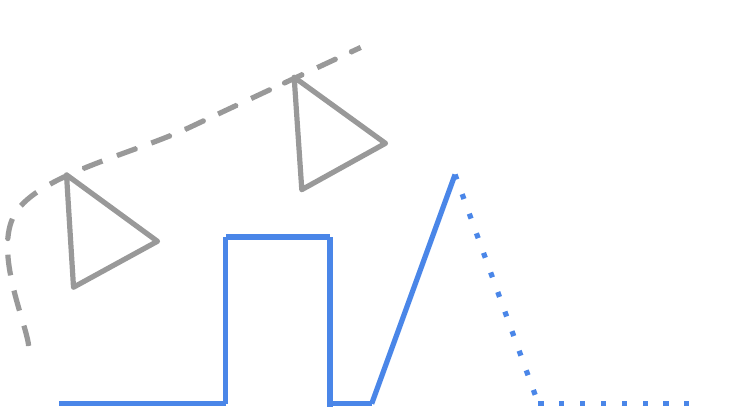}
  &
  \parbox{4cm}{
    (b) As reconstruction algorithms improved, datasets have used videos to capture more of the scene.
    These still miss the backs of many objects. \vspace{4pt} \\
    Example: \cite{xiao-iccv-2013} }
  \\
  \vspace{8pt}

  \includegraphics[width=0.49\columnwidth]{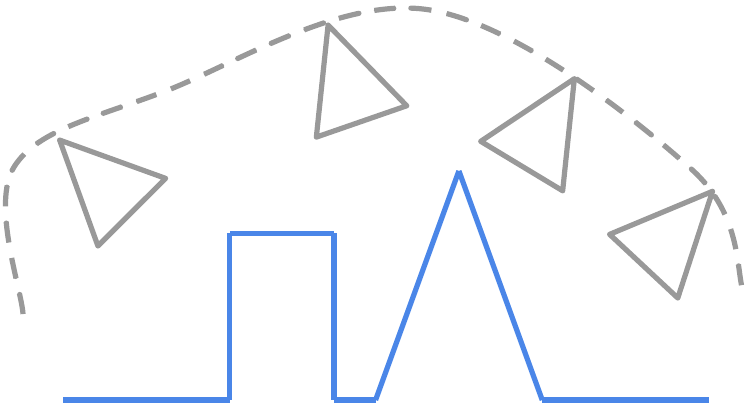}
  &
  \parbox{4cm}{
    (c) Few datasets capture the full visible surface geometry. \vspace{4pt} \\
    Examples: \cite{choi-arxiv-2016} captures objects, and  \cite{firman-cvpr-2016} captures tabletop scenes.}
  \\
  \vspace{8pt}

  \includegraphics[width=0.49\columnwidth]{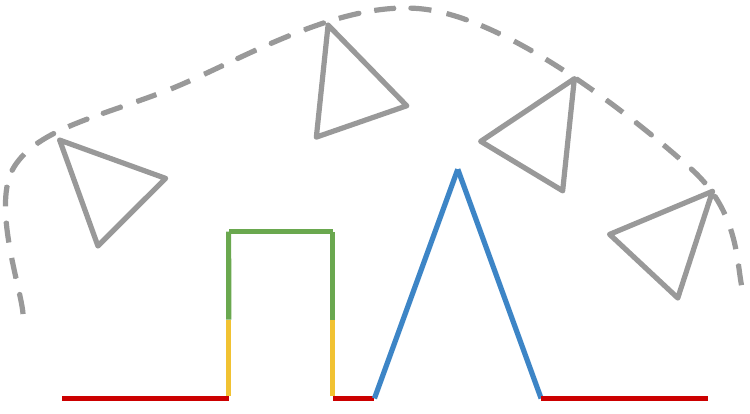}
  &
  \parbox{4cm}{
    (d) No datasets, to our knowledge, capture the full surface geometry of scenes \textit{and} provide semantic labeling on the observed surface. \vspace{4pt}  \\
  }
  \\
  \vspace{8pt}

  \includegraphics[width=0.49\columnwidth]{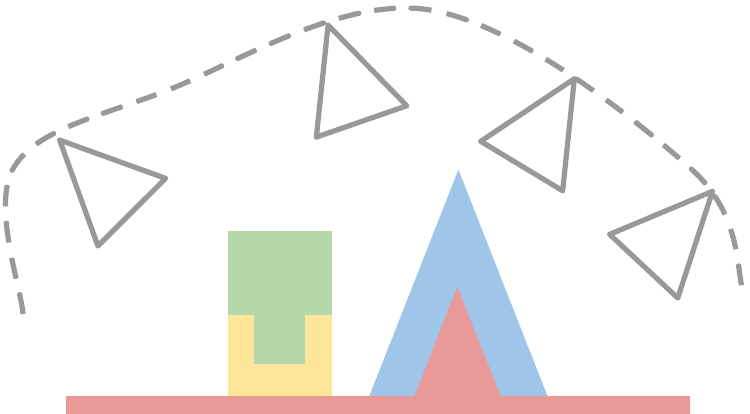}
  &
  \parbox{4cm}{
    (e) Furthermore, our world extends beyond the visible surface. Dense \textit{volumetric} labeling of scenes would enable a deeper level of understanding and interaction. \\
    }
  \\

  \end{tabular}
\vspace{10pt}
\caption{\label{fig:volume} Datasets progress to include more 3D information.}
\end{figure}


Most existing semantic datasets view the world as a 2.5D image, where only surfaces directly viewed from one static camera position are visible (Figure \ref{fig:volume}a).
Even datasets with videos (e.g. SUN3D \cite{xiao-iccv-2013}) tend to fail to capture the full surface geometry of scenes  (Figure \ref{fig:volume}b).
Full surface geometry is captured on an object level by \cite{choi-arxiv-2016} and on tabletop scenes by \cite{firman-cvpr-2016} (Figure \ref{fig:volume}c), but capturing and reconstructing a dataset of large, real-world scenes is left as an open challenge.

Labeling the surfaces of such dense reconstructions (Figure \ref{fig:volume}d) would allow for semantic segmentation on a \emph{mesh} level.
Many opportunities would be afforded by datasets which provide labeled on this form of dense reconstruction rather than on images or videos.

Furthermore, we can imagine the benefits of an algorithm which could segment or semantically label a scene on a \emph{voxel} level, following works such as \cite{kim-iccv-2013}.
To train and validate such a system we would require a dataset containing semantic labeling of each voxel in a scene (Figure~\ref{fig:volume}e).
The difficulty of applying such labeling by hand may make synthetic data necessary for this problem.

\subsection{Geometry of dynamic scenes}

Aside from a single sequence from \cite{varol-cvpr-2012}, we know of no RGBD datasets captured from \emph{dynamic} scenes with ground truth dense geometry.
One option is to use deformable meshes provided for face datasets  \cite{garrido-acm-2013, valgaerts-acm-2012} or fabrics \cite{garg-emmcvpr-2011}, which can be synthetically re-rendered to give dense correspondences between frames (\eg Zollh{\" o}efer \ea \cite{zollhoefer-tog-2014} re-render data from \cite{valgaerts-acm-2012}).
Datasets of humans with motion capture data (Section \ref{sec:actions}) also give a very sparse dense geometry with correspondences.

The open challenge for the field of dense reconstruction is to directly capture an RGBD dataset of deforming objects with ground truth geometry and correspondences between frames.


\section{Conclusion}

We have discovered a considerable quantity of RGBD datasets available for researchers to use.
While some overlap in their scope, overall the field is promisingly diverse which suggests that depth information is useful in many different sectors.

Most datasets we reviewed have been captured as single frames or videos from static cameras.
We are now entering an era where the \emph{collection} and \emph{labeling} of datasets requires state-of-the-art computer vision research.
For example, capturing a dense dataset such as \cite{choi-arxiv-2016} would not have been possible when the Kinect was first launched.
As reconstruction and labeling algorithms for RGBD data improve, the community has a massive opportunity to create and share new datasets of 3D reconstructions of static, and ultimately dynamic scenes.

%
%
%
\clearpage
\section*{Acknowledgements}
I am extremely grateful to Gabriel Brostow and his group for their relentless support, and to Lourdes Agapito for her helpful discussions.
A big thanks also goes out to everyone who has released their datasets.
Keep them coming!
\section*{References}
\noindent For references which refer to a dataset we give a URL to the project page from which the data can be downloaded.

{\fontsize{9}{10}\selectfont
\bibliographystyle{ieee}
\bibliography{bibtex/strings.bib,bibtex/refs_w_url.bib}

\begin{thebibliography}{100}\itemsep=-1pt

\bibitem{aksor-ras-2014}
E.~E. Aksoy, M.~Tamosiunaite, and F.~W{\"o}rg{\"o}tter.
\newblock Model-free incremental learning of the semantics of manipulation
  actions.
\newblock {\em Robotics and Autonomous Systems}, 2014.
\newblock \url{http://www.dpi.physik.uni-goettingen.de/~eaksoye/dataset.html}.

\bibitem{aldoma-www-2012}
A.~Aldoma and A.~Richtsfeld.
\newblock {The Willow Garage Object Recognition Challenge}, 2012.
\newblock
  \url{http://www.acin.tuwien.ac.at/forschung/v4r/mitarbeiterprojekte/willow/}.

\bibitem{aly-icb-2015}
S.~Aly, A.~Trubanova, A.~L. Abbott, S.~W. White, and A.~E. Youssef.
\newblock {VT-KFER: A Kinect-based RGBD}+time dataset for spontaneous and
  non-spontaneous facial expression recognition.
\newblock In {\em International Conference on Biometrics}, 2015.
\newblock \url{http://sufficiency.ece.vt.edu/VT-KFER/}.

\bibitem{savran-cost-2008}
A.~S.~N. Aly{\" u}z, H.~Dibeklio{\v g}lu, O.~{\c C}eliktutan, B.~G{\" o}kberk,
  B.~Sankur, and L.~Akarun.
\newblock Bosphorus database for {3D} face analysis.
\newblock In {\em COST 2101 Workshop on Biometrics and Identity Management
  (BIOID 2008)}, 2008.
\newblock \url{http://bosphorus.ee.boun.edu.tr/Home.aspx}.

\bibitem{amiri-hc-2013}
S.~M. Amiri, M.~T. Pourazad, P.~Nasiopoulos, and V.~C. Leung.
\newblock Non-intrusive human activity monitoring in a smart home environment.
\newblock In {\em Conference on e-Health Networking, Application and Services
  (IEEE Healthcom 2013)}, 2013.
\newblock \url{http://dml.ece.ubc.ca/data/smartaction/}.

\bibitem{varol-cvpr-2012}
A.Varol, M.~Salzmann, P.~Fua, and R.~Urtasun.
\newblock A constrained latent variable model.
\newblock In {\em Computer Vision and Pattern Recognition (CVPR)}, 2012.
\newblock \url{http://cvlab.epfl.ch/data/dsr}.

\bibitem{bagautdinov-cvpr-2015}
T.~Bagautdinov, F.~Fleuret, and P.~Fua.
\newblock Probability occupancy maps for occluded depth images.
\newblock In {\em Computer Vision and Pattern Recognition (CVPR)}, 2015.
\newblock \url{http://cvlab.epfl.ch/data/rgbd-pedestrian}.

\bibitem{barbosa-eccvw-2012}
I.~B. Barbosa, M.~Cristani, A.~D. Bue, L.~Bazzani, and V.~Murino.
\newblock Re-identification with {RGB-D} sensors.
\newblock In {\em First International Workshop on Re-Identification}, 2012.
\newblock \url{http://www.iit.it/en/datasets-and-code/datasets/rgbdid.html}.

\bibitem{berger-arxiv-2013}
K.~Berger.
\newblock The role of {RGB-D} benchmark datasets: an overview.
\newblock {\em arXiv:1310.2053}, 2013.

\bibitem{bloom-eccvw-2014}
V.~Bloom, V.~Argyriou, and D.~Makris.
\newblock {G3Di}: A gaming interaction dataset with a real time detection and
  evaluation framework.
\newblock In {\em European Conference on Computer Vision (ECCV) Workshops},
  2014.
\newblock \url{http://dipersec.king.ac.uk/G3D/}.

\bibitem{bloom-cvprw-2012}
V.~Bloom, D.~Makris, and V.~Argyriou.
\newblock {G3D}: A gaming action dataset and real time action recognition
  evaluation framework.
\newblock In {\em Computer Vision and Pattern Recognition (CVPR) Workshops},
  2012.
\newblock \url{http://dipersec.king.ac.uk/G3D/}.

\bibitem{breitenstein-cvpr-2008}
M.~D. Breitenstein, D.~Kuettel, T.~Weise, L.~van Gool, and H.~Pfister.
\newblock Real-time face pose estimation from single range images.
\newblock In {\em Computer Vision and Pattern Recognition (CVPR)}, 2008.
\newblock \url{http://www.vision.ee.ethz.ch/datasets/headposeCVPR08/}.

\bibitem{bronstein-vcg-2007}
A.~M. Bronstein, M.~M. Bronstein, and R.~Kimmel.
\newblock Calculus of non-rigid surfaces for geometry and texture manipulation.
\newblock {\em Transactions on Visualization and Computer Graphics}, 2007.
\newblock \url{http://tosca.cs.technion.ac.il/book/resources_data.html}.

\bibitem{butler-eccv-2012}
D.~J. Butler, J.~Wulff, G.~B. Stanley, and M.~J. Black.
\newblock A naturalistic open source movie for optical flow evaluation.
\newblock In {\em European Conference on Computer Vision (ECCV)}, 2012.
\newblock \url{http://sintel.is.tue.mpg.de/}.

\bibitem{calli-icar-2015}
B.~Calli, A.~Singh, A.~Walsman, S.~Srinivasa, P.~Abbeel, and A.~Dollar.
\newblock The {YCB} object and model set: Towards common benchmarks for
  manipulation research.
\newblock In {\em Conference on Advanced Robotics (ICAR)}, 2015.
\newblock \url{http://rll.eecs.berkeley.edu/ycb/}.

\bibitem{chen-icip-2015}
C.~Chen, R.~Jafari, and N.~Kehtarnavaz.
\newblock {UTD-MHAD}: A multimodal dataset for human action recognition
  utilizing a depth camera and a wearable inertial sensor.
\newblock In {\em International Conference on Image Processing (ICIP)}, 2015.
\newblock \url{http://www.utdallas.edu/~kehtar/UTD-MHAD.html}.

\bibitem{chen-vcg-2014}
C.~Chen, Y.~Weng, S.~Zhou, Y.~Tong, and K.~Zhou.
\newblock {FaceWarehouse}: a {3D} facial expression database for visual
  computing.
\newblock {\em Transactions on Visualization and Computer Graphics}, 2014.
\newblock \url{http://gaps-zju.org/facewarehouse/}.

\bibitem{cheng-eccvw-2012}
Z.~Cheng, L.~Qin, Y.~Ye, Q.~Huang, and Q.~Tian.
\newblock Human daily action analysis with multi-view and color-depth data.
\newblock In {\em European Conference on Computer Vision (ECCV) workshops},
  2012.
\newblock \url{http://vipl.ict.ac.cn/rgbd-action-dataset}.

\bibitem{casia-minolta-2008}
{Chinese Academy of Sciences' Institute of Automation (CASIA)}.
\newblock {CASIA-3D FaceV1}, 2008.
\newblock \url{http://biometrics.idealtest.org/}.

\bibitem{choi-cvpr-2015}
S.~Choi, Q.-Y. Zhou, and V.~Koltun.
\newblock Robust reconstruction of indoor scenes.
\newblock In {\em Computer Vision and Pattern Recognition (CVPR)}, 2015.
\newblock \url{http://redwood-data.org/indoor/dataset.html}.

\bibitem{choi-arxiv-2016}
S.~Choi, Q.-Y. Zhou, S.~Miller, and V.~Koltun.
\newblock A large dataset of object scans.
\newblock {\em arXiv:1602.02481}, 2016.
\newblock \url{http://redwood-data.org/3dscan}.

\bibitem{cippitelli-icc-2015}
E.~Cippitelli, S.~Gasparrini, E.~Gambi, S.~Spinsante, J.~Wahslen, I.~Orhan, and
  T.~Lindh.
\newblock Time synchronization and data fusion for {RGB-Depth} cameras and
  wearable inertial sensors in {AAL} applications.
\newblock In {\em {ICC2015} - Workshop on {ICT}-Enabled Services and
  Technologies for {eHealth} and Ambient Assisted Living}, 2015.
\newblock \url{http://www.tlc.dii.univpm.it/blog/databases4kinect}.

\bibitem{dib-iros-2015}
A.~Dip and F.~Charpillet.
\newblock Pose estimation for a partially observable human body from {RGB-D}
  cameras.
\newblock In {\em Intelligent Robots and Systems (IROS)}, 2015.
\newblock \url{https://team.inria.fr/larsen/software/datasets/}.

\bibitem{doumanoglou-ivmsp-2013}
A.~Doumanoglou, S.~Asteriadis, D.~S. Alexiadis, D.~Zarpalas, and P.~Daras.
\newblock A dataset of {Kinect}-based {3D} scans.
\newblock In {\em 3D Image/Video Technologies and Applications}, 2013.
\newblock \url{http://vcl.iti.gr/3d-scans/}.

\bibitem{erdogmus-btas-2013}
N.~Erdogmus and S.~Marcel.
\newblock Spoofing in {2D} face recognition with {3D} masks and anti-spoofing
  with {Kinect}.
\newblock {\em Biometrics: Theory, Applications and Systems}, 2013.
\newblock \url{https://www.idiap.ch/dataset/3dmad}.

\bibitem{escalera-eccv-2014}
S.~Escalera, X.~Bar{\'o}, J.~Gonz{\'a}lez, M.~A. Bautista, M.~Madadi, M.~Reyes,
  V.~Ponce-L{\'o}pez, H.~J. Escalante, J.~Shotton, and I.~Guyon.
\newblock {ChaLearn} looking at people challenge 2014: Dataset and results.
\newblock In {\em European Conference on Computer Vision (ECCV) Workshops},
  2014.
\newblock
  \url{http://gesture.chalearn.org/2013-multi-modal-challenge/data-2013-challenge}.

\bibitem{fanelli-tom-2010}
G.~Fanelli, J.~Gall, H.~Romsdorfer, T.Weise, and L.~V. Gool.
\newblock A {3-D} audio-visual corpus of affective communication.
\newblock {\em IEEE Transactions on Multimedia}, 2010.
\newblock
  \url{http://www.vision.ee.ethz.ch/~gfanelli/head_pose/head_forest.html#db}.

\bibitem{fanelli-dadm-2011}
G.~Fanelli, T.~Weise, J.~Gall, and L.~V. Gool.
\newblock Real time head pose estimation from consumer depth cameras.
\newblock In {\em Annual Symposium of the German Association for Pattern
  Recognition (DAGM)}, 2011.
\newblock \url{http://www.vision.ee.ethz.ch/datasets/b3dac2.en.html}.

\bibitem{firman-iros-2011}
M.~Firman and S.~Julier.
\newblock `{Misspelled}' visual words in unsupervised range data
  classification: The effect of noise on classification performance.
\newblock In {\em Intelligent Robots and Systems (IROS)}, 2011.

\bibitem{firman-cvpr-2016}
M.~Firman, O.~Mac~Aodha, S.~Julier, and G.~Brostow.
\newblock Structured prediction of unobserved voxels from a single depth image.
\newblock In {\em Computer Vision and Pattern Recognition (CVPR)}, 2016.
\newblock (To appear). \url{http://visual.cs.ucl.ac.uk/pubs/depthPrediction/}.

\bibitem{firman-iros-2013}
M.~Firman, D.~Thomas, S.~Julier, and A.~Sugimoto.
\newblock Learning to discover objects in {RGB-D} images using correlation
  clustering.
\newblock In {\em Intelligent Robots and Systems (IROS)}, 2013.

\bibitem{fisher-siggraph-2012}
M.~Fisher, D.~Ritchie, M.~Savva, T.~Funkhouser, and P.~Hanrahan.
\newblock Example-based synthesis of 3{D} object arrangements.
\newblock {\em ACM Transactions on Graphics}, 2012.

\bibitem{fothergill-chi-2012}
S.~Fothergill, H.~M. Mentis, P.~Kohli, and S.~Nowozin.
\newblock Instructing people for training gestural interactive systems.
\newblock In {\em Human Factors in Computing Systems (CHI)}, 2012.
\newblock
  \url{http://research.microsoft.com/en-us/um/cambridge/projects/msrc12/}.

\bibitem{freeman-arxiv-2015}
D.~Freeman, R.~Jota, D.~Vogel, D.~Wigdor, and R.~Balakrishnan.
\newblock A dataset of naturally occurring, whole-body background activity to
  reduce gesture conflicts.
\newblock {\em arXiv:1509.06109}, 2015.
\newblock \url{http://www.dgp.toronto.edu/%CB%9Cdustin/backgroundactivity/}.

\bibitem{garg-emmcvpr-2011}
R.~Garg, A.~Roussos, and L.~Agapito.
\newblock Robust trajectory-space {TV-L1} optical flow for non-rigid sequences.
\newblock In {\em Energy Minimization Methods in Computer Vision and Pattern
  Recognition (EMMCVPR)}, 2011.
\newblock \url{http://www0.cs.ucl.ac.uk/staff/lagapito/subspace_flow/}.

\bibitem{garrido-acm-2013}
P.~Garrido, L.~Valgaerts, C.~Wu, and C.~Theobalt.
\newblock Reconstructing detailed dynamic face geometry from monocular video.
\newblock In {\em {ACM} Trans. Graph. (Proceedings of SIGGRAPH Asia 2013)},
  2013.
\newblock \url{http://gvv.mpi-inf.mpg.de/projects/MonFaceCap/}.

\bibitem{gasparrini-techall-2015}
S.~Gasparrini, E.~Cippitelli, E.~Gambi, S.~Spinsante, and F.~F. Revuelta.
\newblock Performance analysis of self-organising neural networks tracking
  algorithms for intake monitoring using {Kinect}.
\newblock In {\em Technologies for Active and Assisted Living (TechAAL)}, 2015.
\newblock \url{http://www.tlc.dii.univpm.it/blog/databases4kinect}.

\bibitem{gasparrini-ictinv-2015}
S.~Gasparrini, E.~Cippitelli, E.~Gambi, S.~Spinsante, J.~Wahslen, I.~Orhan, and
  T.~Lindh.
\newblock Proposal and experimental evaluation of fall detection solution based
  on wearable and depth data fusion.
\newblock In {\em ICT Innovations 2015, Workshop ELEMENT}, 2015.
\newblock \url{http://www.tlc.dii.univpm.it/blog/databases4kinect}.

\bibitem{gasparrini-sensors-2014}
S.~Gasparrini, E.~Cippitelli, S.~Spinsante, and E.~Gambi.
\newblock A depth-based fall detection system using a
  {K}inect{\textsuperscript{\textregistered}} sensor.
\newblock {\em Sensors}, 2014.
\newblock \url{http://www.tlc.dii.univpm.it/blog/databases4kinect}.

\bibitem{gschwandtner-isvc-2011}
M.~Gschwandtner, R.~Kwitt, and A.~Uhl.
\newblock {BlenSor}: Blender sensor simulation toolbox.
\newblock In {\em Advances in Visual Computing: 7th International Symposium,
  (ISVC)}, 2011.

\bibitem{guo-iccv-2013}
R.~Guo and D.~Hoiem.
\newblock Support surface prediction in indoor scenes.
\newblock In {\em International Conference on Computer Vision (ICCV)}, 2013.
\newblock \url{http://aqua.cs.uiuc.edu/site/projects/scenemodel.html}.

\bibitem{guo-iciea-2014}
Y.~Guo, J.~Zhang, M.~Lu, J.~Wan, and Y.~Ma.
\newblock Benchmark datasets for {3D} computer vision.
\newblock In {\em Industrial Electronics and Applications (ICIEA)}, 2014.

\bibitem{handa-arxiv-2015}
A.~Handa, V.~Patraucean, V.~Badrinarayanan, S.~Stent, and R.~Cipolla.
\newblock {SceneNet}: Understanding real world indoor scenes with synthetic
  data.
\newblock {\em arXiv:1601.05511}, 2016.
\newblock \url{http://robotvault.bitbucket.org/}.

\bibitem{handa-icra-2014}
A.~Handa, T.~Whelan, J.~McDonald, and A.~J. Davison.
\newblock A benchmark for {RGB-D} visual odometry, {3D} reconstruction and
  {SLAM}.
\newblock In {\em International Conference on Robotics and Automation (ICRA)},
  2014.
\newblock \url{http://www.doc.ic.ac.uk/~ahanda/VaFRIC/iclnuim.html}.

\bibitem{hinterstoisser-accv-2012}
S.~Hinterstoisser, V.~Lepetit, S.~Ilic, S.~Holzer, G.~Bradski, K.~Konolige, and
  N.~Navab.
\newblock Model based training, detection and pose estimation of texture-less
  {3D} objects in heavily cluttered scenes.
\newblock In {\em Asian Conference on Computer Vision (ACCV)}, 2012.
\newblock \url{http://campar.in.tum.de/Main/StefanHinterstoisser}.

\bibitem{hog-sitibs-2012}
R.~H{\o}g, P.~Jasek, C.~Rofidal, K.~Nasrollahi, T.~Moeslund, and G.~Tranchet.
\newblock An {RGB-D} database using {Microsoft’s Kinect for Windows} for face
  detection.
\newblock In {\em International Conference on Signal Image Technology \&
  Internet Based Systems}, 2012.
\newblock \url{http://www.vap.aau.dk/rgb-d-face-database/}.

\bibitem{hsiao-msc-2014}
Y.~Hsiao, J.~Sanchez{-}Riera, T.~Lim, K.~Hua, and W.~Cheng.
\newblock {LaRED}: a large {RGB-D} extensible hand gesture dataset.
\newblock In {\em Multimedia Systems Conference}, 2014.
\newblock \url{http://mclab.citi.sinica.edu.tw/dataset/lared/lared.html}.

\bibitem{hu-mpe-2013}
T.~Hu, X.~Zhu, W.~Guo, and K.~Su.
\newblock Efficient interactions recognition through positive action based
  representation.
\newblock {\em Mathematical Problems in Engineering}, 2013.
\newblock
  \url{http://www.lmars.whu.edu.cn/prof_web/zhuxinyan/DataSetPublish/dataset.html}.

\bibitem{guyon-adiaa-2012}
I.Guyon, V.~Athitsos, P.~Jangyodsuk, H.~J. Escalante, and B.~Hamner.
\newblock Results and analysis of the {ChaLearn} gesture challenge 2012.
\newblock In {\em Advances in Depth Image Analysis and Applications}, 2012.
\newblock \url{http://gesture.chalearn.org/data}.

\bibitem{ikkala-visapp-2016}
A.~Ikkala, J.~Pajarinen, and V.~Kyrki.
\newblock Benchmarking {RGB-D} segmentation: Toy dataset of complex crowded
  scenes.
\newblock In {\em Computer Vision Theory and Applications (VISAPP)}, 2016.
\newblock \url{http://irobotics.aalto.fi/software-and-data/toy-dataset}.

\bibitem{ionescu-pami-2014}
C.~Ionescu, D.~Papava, V.~Olaru, and C.~Sminchisescu.
\newblock Human3.6m: Large scale datasets and predictive methods for {3D} human
  sensing in natural environments.
\newblock {\em Pattern Analysis and Machine Intelligence (PAMI)}, 2014.
\newblock \url{http://vision.imar.ro/human3.6m}.

\bibitem{janoch-iccv-2011}
A.~Janoch, S.~Karayev, Y.~Jia, J.~T. Barron, M.~Fritz, K.~Saenko, and
  T.~Darrell.
\newblock A category-level {3-D} object dataset: Putting the {Kinect} to work.
\newblock In {\em International Conference on Computer Vision (ICCV)Workshop on
  Consumer Depth Cameras in Computer Vision}, 2011.
\newblock \url{http://kinectdata.com/}.

\bibitem{karg-aamas-2014}
M.~Karg and A.~Kirsch.
\newblock A human morning routine dataset.
\newblock In {\em Conference on Autonomous Agents and Multiagent Systems
  (AAMAS)}, 2014.
\newblock {Extended Abstract} \url{http://tinyurl.com/zzsbr4j}.

\bibitem{kasper-ijrr-2012}
A.~Kasper, Z.~Xue, and R.~Dillmann.
\newblock The {KIT} object models database: An object model database for object
  recognition, localization and manipulation in service robotics.
\newblock {\em International Journal of Robotics Research}, 2012.
\newblock \url{http://his.anthropomatik.kit.edu/objectmodels/} [Link broken as
  of March 2016].

\bibitem{kastaniotis-prl-2015}
D.~Kastaniotis, I.~Theodorakopoulos, C.~Theoharatos, G.~Economou, and
  S.~Fotopoulos.
\newblock A framework for gait-based recognition using {Kinect}.
\newblock {\em Pattern Recognition Letters}, 2015.
\newblock \url{http://www.upcv.upatras.gr/personal/kastaniotis/datasets.html}.

\bibitem{kim-iccv-2013}
B.-s. Kim, P.~Kohli, and S.~Savarese.
\newblock {3D} scene understanding by voxel-{CRF}.
\newblock In {\em International Conference on Computer Vision (ICCV)}, 2013.

\bibitem{koppula-nips-2011}
H.~S. Koppula, A.~Anand, T.~Joachims, and A.~Saxena.
\newblock Semantic labeling of {3D} point clouds for indoor scenes.
\newblock In {\em Neural Information Processing (NIPS)}, 2011.
\newblock \url{http://pr.cs.cornell.edu/sceneunderstanding/data/data.php}.

\bibitem{koppula-ijrr-2013}
H.~S. Koppula, R.~Gupta, and A.~Saxena.
\newblock Learning human activities and object affordances from {RGB-D} videos.
\newblock {\em International Journal of Robotics Research (IJRR)}, 2013.
\newblock \url{http://pr.cs.cornell.edu/humanactivities/data.php}.

\bibitem{kurakin-eusipco-2012}
A.~Kurakin, Z.~Zhang, and Z.~Liu.
\newblock A real-time system for dynamic hand gesture recognition with a depth
  sensor.
\newblock In {\em EUSIPCO}, 2012.
\newblock
  \url{http://research.microsoft.com/en-us/um/people/zliu/actionrecorsrc/}.

\bibitem{kwolek-cmpb-2014}
B.~Kwolek and M.~Kepski.
\newblock Human fall detection on embedded platform using depth maps and
  wireless accelerometer.
\newblock {\em Computer Methods and Programs in Biomedicine}, 2014.
\newblock \url{http://fenix.univ.rzeszow.pl/~mkepski/ds/uf.html}.

\bibitem{lai-icra-2014}
K.~Lai, L.~Bo, and D.~Fox.
\newblock Unsupervised feature learning for {3D} scene labeling.
\newblock In {\em International Conference on Robotics and Automation (ICRA)},
  2014.
\newblock \url{http://rgbd-dataset.cs.washington.edu/dataset/rgbd-scenes-v2/}.

\bibitem{lai-icra-2011}
K.~Lai, L.~Bo, X.~Ren, and D.~Fox.
\newblock A large-scale hierarchical multi-view {RGB-D} object dataset.
\newblock In {\em International Conference on Robotics and Automation (ICRA)},
  2011.
\newblock \url{http://rgbd-dataset.cs.washington.edu/dataset/rgbd-scenes/}.

\bibitem{lai-icra-2012}
K.~Lai, L.~Bo, X.~Ren, and D.~Fox.
\newblock Detection-based object labelling in {3D} scenes.
\newblock In {\em International Conference on Robotics and Automation (ICRA)},
  2012.

\bibitem{leightley-apsipa-2015}
D.~Leightley, M.~Yap, J.~Coulson, Y.~Barnouin, and J.~McPhee.
\newblock Benchmarking human motion analysis using {Kinect One}: an open source
  dataset.
\newblock In {\em ,IEEE International Conference by Asia-Pacific Signal and
  Information Processing Association}, 2015.
\newblock \url{http://k3da.leightley.com/}.

\bibitem{li-cvprw-2010}
W.~Li, Z.~Zhang, and Z.~Liu.
\newblock Action recognition based on a bag of {3D} points.
\newblock In {\em Workshop on CVPR for Human Communicative Behavior Analysis},
  2010.
\newblock
  \url{http://research.microsoft.com/en-us/um/people/zliu/actionrecorsrc/}.

\bibitem{lillo-cvpr-2014}
I.~Lillo, A.~Soto, and J.~C. Niebles.
\newblock Discriminative hierarchical modeling of spatio-temporally composable
  human activities.
\newblock In {\em Computer Vision and Pattern Recognition (CVPR)}, 2014.
\newblock \url{http://web.ing.puc.cl/~ialillo/ActionsCVPR2014/}.

\bibitem{liu-sp-2015}
A.~Liu, W.~Nie, Y.~Su, L.~Ma, T.~Hao, and Z.~Yang.
\newblock Coupled hidden conditional random fields for {RGB-D} human action
  recognition.
\newblock {\em Signal Processing}, 2015.
\newblock \url{http://media.tju.edu.cn/tju_dataset.html}.

\bibitem{liu-is-2014}
A.~Liu, Z.~Wang, W.~Nie, and Y.~Su.
\newblock Graph-based characteristic view set extraction and matching for {3D}
  model retrieval.
\newblock {\em Information Sciences}, 2015.
\newblock \url{http://media.tju.edu.cn/mvred/dataset1.html}. Under review.

\bibitem{liu-ijcai-2013}
L.~Liu and L.~Shao.
\newblock Learning discriminative representations from {RGB-D} video data.
\newblock In {\em International Joint Conference on Artificial Intelligence},
  2013.
\newblock \url{http://lshao.staff.shef.ac.uk/data/SheffieldKinectGesture.htm}.

\bibitem{luber-iros-2011}
M.~Luber, L.~Spinello, and K.~O. Arras.
\newblock People tracking in {RGB-D} data with on-line boosted target models.
\newblock In {\em Intelligent Robots and Systems (IROS)}, 2011.
\newblock
  \url{http://www2.informatik.uni-freiburg.de/~spinello/RGBD-dataset.html}.

\bibitem{marin-icip-2014}
G.~Marin, F.~Dominio, and P.~Zanuttigh.
\newblock Hand gesture recognition with leap motion and {Kinect} devices.
\newblock In {\em International Conference on Image Processing (ICIP)}, 2014.
\newblock \url{http://lttm.dei.unipd.it/downloads/gesture/index.html}.

\bibitem{martinezgomez-ijrr-2015}
J.~Mart{\'i}nez-G{\'o}mez, I.~Garc{\'i}a-Varea, M.~Cazorla, and V.~Morell.
\newblock {ViDRILO}: The visual and depth robot indoor localization with
  objects information dataset.
\newblock {\em International Journal of Robotics Research}, 2015.
\newblock \url{http://www.rovit.ua.es/dataset/vidrilo/}.

\bibitem{mason-iros-2012}
J.~Mason, B.~Marthi, and R.~Parr.
\newblock Object disappearance for object discovery.
\newblock In {\em Intelligent Robots and Systems (IROS)}, 2012.
\newblock \url{http://wiki.ros.org/Papers/IROS2012_Mason_Marthi_Parr}.

\bibitem{meister-iros-2012}
S.~Meister, S.~Izadi, P.~Kohli, M.~H{\"a}mmerle, C.~Rother, and D.~Kondermann.
\newblock When can we use {KinectFusion} for ground truth acquisition?
\newblock In {\em Intelligent Robots and Systems (IROS) Workshop on Color-Depth
  Camera Fusion in Robotics}, 2012.
\newblock
  \url{http://hci.iwr.uni-heidelberg.de//Benchmarks/document/kinectFusionCapture/}.

\bibitem{meister-vmv-2013}
S.~Meister, R.~Nair, and D.~Kondermann.
\newblock Simulation of time-of-flight sensors using global illumination.
\newblock In {\em Vision, Modeling, and Visualization Workshop}, 2013.

\bibitem{memo-stag-2015}
A.~Memo, L.~Minto, and P.~Zanuttigh.
\newblock Exploiting silhouette descriptors and synthetic data for hand gesture
  recognition.
\newblock In {\em STAG: Smart Tools \& Apps for Graphics}, 2015.
\newblock \url{http://lttm.dei.unipd.it/downloads/gesture/index.html}.

\bibitem{mian-pami-2006}
A.~Mian, M.~Bennamoun, and R.~Owens.
\newblock {3D} model-bsed object recognition and segmentation in cluttered
  scenes.
\newblock {\em Pattern Analysis and Machine Intelligence (PAMI)}, 2006.
\newblock \url{http://www.csse.uwa.edu.au/~ajmal/recognition.html}.

\bibitem{min-smc-2014}
R.~Min, N.~Kose, and J.-L. Dugelay.
\newblock {KinectFaceDB}: A {Kinect} database for face recognition.
\newblock {\em IEEE Transactions on Systems, Man, and Cybernetics}, 2014.
\newblock \url{http://rgb-d.eurecom.fr/}.

\bibitem{moreels-iccv-2005}
P.~Moreels and P.~Perona.
\newblock Evaluation of features detectors and descriptors based on {3D}
  objects.
\newblock In {\em International Conference on Computer Vision (ICCV)}, 2005.

\bibitem{munaro-icra-2014}
M.~Munaro, A.~Basso, A.~Fossati, L.~V. Gool, and E.~Menegatti.
\newblock {3D} reconstruction of freely moving persons for re-identification
  with a depth sensor.
\newblock In {\em International Conference on Robotics and Automation (ICRA)},
  2014.
\newblock
  \url{http://robotics.dei.unipd.it/reid/index.php/8-dataset/5-overview-iaslab}.

\bibitem{munaro-iros-2012}
M.~Munaro, F.~Basso, and E.~Menegatti.
\newblock People tracking within groups with {RGB-D} data.
\newblock In {\em Intelligent Robots and Systems (IROS)}, 2012.
\newblock \url{http://www.dei.unipd.it/~munaro/KTP-dataset.html}.

\bibitem{ni-iccvw-2011}
B.~Ni, G.~Wang, and P.~Moulin.
\newblock {RGBD-HuDaAct}: A color-depth video database for human daily activity
  recognition.
\newblock In {\em IEEE Workshop on Consumer Depth Cameras for Computer Vision
  in conjunction with ICCV}, 2011.
\newblock
  \url{http://adsc.illinois.edu/sites/default/files/files/ADSC-RGBD-dataset-download-instructions.pdf}.

\bibitem{ofli-wacv-2013}
F.~Ofli, R.~Chaudhry, G.~Kurillo, R.~Vidal, and R.~Bajcsy.
\newblock {Berkeley MHAD}: A comprehensive multimodal human action database.
\newblock In {\em Winter Conference on Applications of Computer Vision (WACV)},
  2013.
\newblock \url{http://tele-immersion.citris-uc.org/berkeley_mhad}.

\bibitem{pomerleau-iros-2011}
F.~Pomerleau, S.~Magnenat, F.~Colas, M.~Liu, and R.~Siegwart.
\newblock Tracking a depth camera: Parameter exploration for fast {ICP}.
\newblock In {\em Intelligent Robots and Systems (IROS)}, 2011.
\newblock
  \url{http://projects.asl.ethz.ch/datasets/doku.php?id=Kinect:iros2011Kinect}.

\bibitem{rennie-ral-2016}
C.~Rennie, R.~Shome, K.~E. Bekris, and A.~F.~D. Souza.
\newblock A dataset for improved {RGBD}-based object detection and pose
  estimation for warehouse pick-and-place.
\newblock {\em IEEE Robotics and Automation Letters}, 2016.
\newblock \url{http://www.pracsyslab.org/rutgers_apc_rgbd_dataset}.

\bibitem{richtsfeld-iros-2012}
A.~Richtsfeld, T.~M{\"o}rwald, J.~Prankl, M.~Zillich, and M.~Vincze.
\newblock Segmentation of unknown objects in indoor environments.
\newblock In {\em Intelligent Robots and Systems (IROS)}, 2012.
\newblock \url{http://www.acin.tuwien.ac.at/?id=289}.

\bibitem{salti-cviu-2014}
S.~Salti, F.~Tombari, and L.~D. Stefano.
\newblock {SHOT}: Unique signatures of histograms for surface and texture
  description.
\newblock {\em Computer Vision and Image Understanding}, 2014.
\newblock \url{http://www.vision.deis.unibo.it/research/80-shot}.

\bibitem{schmidt-acivs-2013}
A.~Schmidt, M.~Fularz, M.~Kraft, A.~Kasiński, and M.~Nowicki.
\newblock An indoor {RGB-D} dataset for the evaluation of robot navigation
  algorithms.
\newblock In {\em Advanced Concepts for Intelligent Vision Systems}. Springer,
  2013.
\newblock \url{http://www.vision.put.poznan.pl/?p=70}.

\bibitem{seidenari-cvprw-2013}
L.~Seidenari, V.~Varano, S.~Berretti, A.~D. Bimbo, and P.~Pala.
\newblock Recognizing actions from depth cameras as weakly aligned multi-part
  bag-of-poses.
\newblock In {\em Computer Vision and Pattern Recognition (CVPR) workshops},
  2013.
\newblock
  \url{http://www.micc.unifi.it/resources/datasets/florence-3d-actions-dataset/}.

\bibitem{shahroudy-cvpr-2016}
A.~Shahroudy, J.~Liu, T.-T. Ng, and G.~Wang.
\newblock {NTU RGB+D}: A large scale dataset for {3D} human activity analysis.
\newblock In {\em Computer Vision and Pattern Recognition (CVPR)}, 2016.

\bibitem{shotton-cvpr-2013}
J.~Shotton, B.~Glocker, C.~Zach, S.~Izadi, A.~Criminisi, and A.~Fitzgibbon.
\newblock Scene coordinate regression forests for camera relocalization in
  {RGB-D} images.
\newblock In {\em Computer Vision and Pattern Recognition (CVPR)}, 2013.
\newblock \url{http://research.microsoft.com/en-us/projects/7-scenes/}.

\bibitem{silberman-iccv-2011}
N.~Silberman and R.~Fergus.
\newblock Indoor scene segmentation using a structured light sensor.
\newblock In {\em International Conference on Computer Vision (ICCV)
  Workshops}, 2011.
\newblock \url{http://cs.nyu.edu/~silberman/datasets/nyu_depth_v1.html}.

\bibitem{silberman-eccv-2012}
N.~Silberman, D.~Hoiem, P.~Kohli, and R.~Fergus.
\newblock Indoor segmentation and support inference from {RGBD} images.
\newblock In {\em European Conference on Computer Vision (ECCV)}, 2012.
\newblock \url{http://cs.nyu.edu/~silberman/datasets/nyu_depth_v2.html}.

\bibitem{singh-icra-2014}
A.~Singh, J.~Sha, K.~Narayan, T.~Achim, and P.~Abbeel.
\newblock {BigBIRD}: A large-scale {3D} database of object instances.
\newblock In {\em International Conference on Robotics and Automation (ICRA)},
  2014.
\newblock \url{http://rll.berkeley.edu/bigbird/}.

\bibitem{song-cvpr-2015}
S.~Song, S.~P. Lichtenberg, and J.~Xiao.
\newblock {SUN RGB-D}: A {RGB-D} scene understanding benchmark suite.
\newblock In {\em Computer Vision and Pattern Recognition (CVPR)}, 2015.
\newblock \url{http://rgbd.cs.princeton.edu/}.

\bibitem{song-iccv-2013}
S.~Song and J.~Xiao.
\newblock Tracking revisited using {RGBD} camera: Unified benchmark and
  baselines.
\newblock In {\em International Conference on Computer Vision (ICCV)}, 2013.
\newblock \url{http://tracking.cs.princeton.edu/dataset.html}.

\bibitem{spinello-iros-2011}
L.~Spinello and K.~O. Arras.
\newblock People detection in {RGB-D} data.
\newblock In {\em Intelligent Robots and Systems (IROS)}, 2011.
\newblock
  \url{http://www2.informatik.uni-freiburg.de/~spinello/RGBD-dataset.html}.

\bibitem{stein-ubicomp-2013}
S.~Stein and S.~J. McKenna.
\newblock Combining embedded accelerometers with computer vision for
  recognizing food preparation activities.
\newblock In {\em UbiComp}, 2013.
\newblock
  \url{http://cvip.computing.dundee.ac.uk/datasets/foodpreparation/50salads/}.

\bibitem{sturm-iros-2012}
J.~Sturm, N.~Engelhard, F.~Endres, W.~Burgard, and D.~Cremers.
\newblock A benchmark for the evaluation of {RGB-D SLAM} systems.
\newblock In {\em Intelligent Robots and Systems (IROS)}, 2012.
\newblock \url{http://vision.in.tum.de/data/datasets/rgbd-dataset}.

\bibitem{sung-pair-2011}
J.~Sung, C.~Ponce, B.~Selman, and A.~Saxena.
\newblock Human activity detection from {RGBD} images.
\newblock In {\em AAAI workshop on Pattern, Activity and Intent Recognition
  (PAIR)}, 2011.
\newblock \url{http://pr.cs.cornell.edu/humanactivities/data.php}.

\bibitem{wandi-eccv-2012}
W.~Susanto, M.~Rohrbach, and B.~Schiele.
\newblock {3D} object detection with multiple {Kinects}.
\newblock In {\em European Conference on Computer Vision (ECCV)}, 2012.
\newblock \url{http://tinyurl.com/hlmwga7}.

\bibitem{theodorakopoulos-jvcir-2013}
I.~Theodorakopoulos, D.~Kastaniotis, G.~Economou, and S.~Fotopoulos.
\newblock Pose-based human action recognition via sparse representation in
  dissimilarity space.
\newblock {\em J. Vis. Commun. Image R.}, 2013.
\newblock \url{http://www.upcv.upatras.gr/personal/kastaniotis/datasets.html}.

\bibitem{tombari-iros-2011}
F.~Tombari, L.~D. Stefano, and S.~Giardino.
\newblock Online learning for automatic segmentation of {3D} data.
\newblock In {\em Intelligent Robots and Systems (IROS)}, 2011.
\newblock \url{http://vision.deis.unibo.it/fede/kinectDataset.html}.

\bibitem{valgaerts-acm-2012}
L.~Valgaerts, C.~Wu, A.~Bruhn, H.-P. Seidel, and C.~Theobalt.
\newblock Lightweight binocular facial performance capture under uncontrolled
  lighting.
\newblock In {\em ACM Transactions on Graphics (Proceedings of SIGGRAPH Asia
  2012)}, 2012.

\bibitem{gemeren-eccv-2014}
C.~van Gemeren, R.~T. Tan, R.~Poppe, and R.~C. Veltkamp.
\newblock Dyadic interaction detection from pose and flow.
\newblock In {\em European Conference on Computer Vision (ECCV)}, 2014.
\newblock \url{http://www.projects.science.uu.nl/shakefive/}.

\bibitem{wang-cvpr-2012}
J.~Wang, Z.~Liu, Y.~Wu, and J.~Yuan.
\newblock Mining actionlet ensemble for action recognition with depth cameras.
\newblock In {\em Computer Vision and Pattern Recognition (CVPR)}, 2012.
\newblock
  \url{http://research.microsoft.com/en-us/um/people/zliu/actionrecorsrc/}.

\bibitem{wang-cvpr-2014}
J.~Wang, X.~Nie, Y.~Xia, Y.~Wu, and S.-C. Zhu.
\newblock Cross-view action modeling, learning and recognition.
\newblock In {\em Computer Vision and Pattern Recognition (CVPR)}, 2014.
\newblock \url{http://users.eecs.northwestern.edu/~jwa368/my_data.html}.

\bibitem{wang-icm-2014}
K.~Wang, X.~Wang, L.~Lin, M.~Wang, and W.~Zuo.
\newblock {3D} human activity recognition with reconfigurable convolutional
  neural networks.
\newblock In {\em ACM International Conference on Multimedia}, 2014.
\newblock \url{http://vision.sysu.edu.cn/projects/3d-activity/}.

\bibitem{wasenmueller-wacv-2016}
O.~Wasenm\"uller, M.~Meyer, and D.~Stricker.
\newblock {CoRBS}: Comprehensive {RGB-D} benchmark for {SLAM} using {Kinect
  v2}.
\newblock In {\em Winter Conference on Applications of Computer Vision (WACV)},
  2016.
\newblock \url{http://corbs.dfki.uni-kl.de/}.

\bibitem{wu-cvpr-2015}
C.~Wu, J.~Zhang, S.~Savarese, and A.~Saxena.
\newblock {Watch-n-Patch}: Unsupervised understanding of actions and relations.
\newblock In {\em Computer Vision and Pattern Recognition (CVPR)}, 2015.
\newblock \url{http://watchnpatch.cs.cornell.edu/}.

\bibitem{xia-cvprw-2012}
L.~Xia, C.~Chen, and J.~Aggarwal.
\newblock View invariant human action recognition using histograms of 3d
  joints.
\newblock In {\em Computer Vision and Pattern Recognition (CVPR) workshops},
  2012.
\newblock \url{http://cvrc.ece.utexas.edu/KinectDatasets/HOJ3D.html}.

\bibitem{xiao-iccv-2013}
J.~Xiao, A.~Owens, and A.~Torralba.
\newblock {SUN3D}: A database of big spaces reconstructed using {SfM} and
  object labels.
\newblock In {\em International Conference on Computer Vision (ICCV)}, 2013.
\newblock \url{http://sun3d.cs.princeton.edu/}.

\bibitem{xu-icm-2015}
N.~Xu, A.~Liu, W.~Nie, Y.~Wong, F.~Li, and Y.~Su.
\newblock Multi-modal \& multi-view \& interactive benchmark dataset for human
  action recognition.
\newblock In {\em ACM International Conference on Multimedia}, 2015.
\newblock \url{http://media.tju.edu.cn/m2i.html}.

\bibitem{yang-cc-2013}
Z.~Yang, L.~Zicheng, and C.~Hong.
\newblock {RGB}-depth feature for {3D} human activity recognition.
\newblock {\em China Communications}, 2013.
\newblock \url{http://www.uestcrobot.net/en/?q=download}.

\bibitem{yu-accv-2014}
G.~Yu, Z.~Liu, and J.~Yuan.
\newblock Discriminative orderlet mining for real-time recognition of
  human-object interaction.
\newblock In {\em Asian Conference on Computer Vision (ACCV)}, 2014.
\newblock
  \url{http://research.microsoft.com/en-us/um/people/zliu/actionrecorsrc/}.

\bibitem{yun-cvprw-2012}
K.~Yun, J.~Honorio, D.~Chattopadhyay, T.~L. Berg, and D.~Samaras.
\newblock Two-person interaction detection using body-pose features and
  multiple instance learning.
\newblock In {\em Computer Vision and Pattern Recognition (CVPR) workshops},
  2012.
\newblock
  \url{http://www3.cs.stonybrook.edu/~kyun/research/kinect_interaction/}.

\bibitem{zhang-arxiv-2016}
J.~Zhang, W.~Li, P.~O. Ogunbona, P.~Wang, and C.~Tanga.
\newblock {RGB-D}-based action recognition datasets: A survey.
\newblock {\em arXiv:1511.07041v2}, 2015.
\newblock \url{http://robotvault.bitbucket.org/}.

\bibitem{zhang-siggraph-2004}
L.~Zhang, N.~Snavely, B.~Curless, and S.~M. Seitz.
\newblock {Spacetime Faces}: High-resolution capture for modeling and
  animation.
\newblock In {\em ACM SIGGRAPH Proceedings}, 2004.
\newblock \url{http://grail.cs.washington.edu/projects/stfaces/}.

\bibitem{zhang-cvpr-2013}
Q.~Zhang, X.~Song, X.~Shao, H.~Zhao, and R.~Shibasaki.
\newblock Category modeling from just a single labeling: Use depth information
  to guide the learning of {2D} models.
\newblock In {\em Computer Vision and Pattern Recognition (CVPR)}, 2013.
\newblock \url{http://shiba.iis.u-tokyo.ac.jp/song/?page_id=343}.

\bibitem{zhou-siggraph-2013}
Q.-Y. Zhou and V.~Koltun.
\newblock Dense scene reconstruction with points of interest.
\newblock {\em ACM Transactions on Graphics}, 2013.
\newblock \url{http://qianyi.info/scenedata.html}.

\bibitem{zollhoefer-tog-2014}
M.~Zollh{\"o}fer, M.~Nie{\ss}ner, S.~Izadi, C.~Rhemann, C.~Zach, M.~Fisher,
  C.~Wu, A.~Fitzgibbon, C.~Loop, C.~Theobalt, and M.~Stamminger.
\newblock Real-time non-rigid reconstruction using an {RGB-D} camera.
\newblock {\em ACM Transactions on Graphics (TOG)}, 2014.

\end{thebibliography}
%
}

\end{document}